\documentclass{article}
\usepackage{log_2026}						

\usepackage{booktabs}						
\usepackage{multirow}						
\usepackage{amsfonts}						
\usepackage{graphicx}						
\usepackage{duckuments}						

\usepackage[numbers,compress,sort]{natbib}	

\usepackage{booktabs}
\usepackage{tabularx}
\usepackage{makecell}
\newcommand{\yes}{\(\checkmark\)}
\newcommand{\no}{\(\times\)}
\newcolumntype{L}[1]{>{\raggedright\arraybackslash}p{#1}}

\usepackage{mathtools}
\usepackage{wrapfig}

\title[Hyperbolic Graph Embedders for Link Prediction and Topology Reconstruction]{Hyperbolic Graph Embedders \\for Link Prediction and Topology Reconstruction}

\author[Jankowski et al.]{%
Robert Jankowski\\
TU Delft \\
\email{R.Jankowski@tudelft.nl}\And
Maksim Kitsak\\
Indiana University\\
\email{maksim.kitsak@gmail.com}\And
Dorota Celińska-Kopczyńska\\
Institute of Informatics, University of Warsaw\\
\email{dot@mimuw.edu.pl}
}

\begin{document}

\maketitle

\begin{abstract}
Hyperbolic embeddings provide compact geometric representations of complex networks in hyperbolic spaces, but systematic comparisons of methods developed in machine learning, network science, and algorithmics remain rare. We benchmark 13 unsupervised hyperbolic graph embedders under a unified protocol for link prediction and topology reconstruction on synthetic and empirical networks. The protocol captures both missing-link recovery and the preservation of local and global network structure. Maximum-likelihood and representation-learning-based approaches, including hybrid variants, achieve the strongest overall performance, although no method dominates across all tasks and structural regimes. Performance is more strongly associated with embedding paradigm than with disciplinary origin. We show that the relative performance of embedding approaches varies across network regimes, highlighting that method selection depends on both the downstream task and the structure of the input network.
\end{abstract}


\section{Introduction}
Networks provide a common representation for relational data across science and engineering, including protein interactions~\cite{rolland2014proteome}, brain connectomes~\cite{sporns2005human}, citation graphs~\cite{Price1965}, communication systems~\cite{van2009performance}, and the World Wide Web~\cite{serrano2003topology}. Despite their different origins, many real networks share structural properties such as heterogeneous degree distributions, short path lengths, and high clustering coefficients~\cite{newman2003structure}.
Hyperbolic geometry offers a natural latent representation of these structures because its exponential volume growth accommodates hierarchical and heterogeneous connectivity patterns.
This insight enabled the development of network models based on hyperbolic geometry~\cite{boguna2021network,krioukov2010hyperbolic,serrano2008self}. In these models, nodes are embedded in hyperbolic space, and links are sampled with a probability that decreases with hyperbolic distance. Consequently, nearby nodes are more likely to be connected than distant ones. These models can generate networks that simultaneously exhibit power-law degree distributions, high clustering coefficients, and the small-world property.

Because many real-world networks exhibit these structural properties, a natural question is whether their underlying hyperbolic geometry can be recovered from the observed topology. This inference task, commonly referred to as hyperbolic network embedding, assigns each node a coordinate in hyperbolic space such that the geometric relationships among nodes capture relevant features of the network’s topology. 
Methods for solving this problem have emerged from several research communities. Machine-learning methods typically optimize proximity-preserving objectives using negative sampling and Riemannian optimization~\cite{nickel2017poincare,nickel2018lorentz}. Network-science methods usually infer coordinates by fitting an explicit geometric random graph model to the observed topology~\cite{papadopoulos2015network,garcia2019mercator}. Algorithmic methods use efficient approximation schemes and graph-based heuristics to infer hyperbolic coordinates~\cite{blasius2016bfkl}. These categories may overlap, but they differ in their assumptions, optimization objectives, and interpretation of the resulting coordinates. We study fully unsupervised embedding methods that use network topology alone, rather than supervised or task-specific graph neural networks. 

We evaluate the inferred coordinates on two complementary tasks: link prediction and topology reconstruction. These tasks are important because they assess whether an embedding can both infer unobserved relationships and capture the structural mechanisms underlying a network, with applications including recommendation systems, social-network analysis, and biological interaction discovery. In link prediction, part of the observed edge set is removed, and candidate node pairs are ranked using hyperbolic distance or an associated connection probability~\cite{kitsak2020link}. This task measures whether the embedding preserves local, edge-level information. In topology reconstruction, the inferred coordinates and model parameters are used to generate synthetic networks. Reconstruction quality is assessed by comparing structural statistics of the generated and original networks, including degree distributions, clustering spectra, degree correlations, and spectral properties. Unlike link prediction, topology reconstruction evaluates whether the embedding provides a faithful generative description of the network. The two tasks therefore capture complementary edge-level and global aspects of embedding quality.



To enable a fair comparison across methods developed in largely separate research communities, we establish a unified and reproducible experimental framework. Our main contributions are:
\begin{itemize}
    \item a benchmark of 13 representative hyperbolic embedding methods from machine learning, network science, and algorithms under a common evaluation protocol;
    \item a joint assessment of edge-level link prediction and graph-level topology reconstruction, which capture complementary aspects of embedding quality;
    \item a systematic investigation of how embedding performance varies with network structure across controlled synthetic networks and diverse empirical networks; and
    \item practical guidance on the structural regimes and evaluation tasks for which different embedding paradigms are most suitable.
\end{itemize}

\section{Theoretical Background}
\paragraph{Hyperbolic geometry} Here, we provide a concise description of hyperbolic geometry. For a more formal treatment, see~\cite{cannon}. Hyperbolic geometry is a non-Euclidean geometry of constant negative curvature, in which initially nearby geodesics diverge exponentially. The hyperbolic plane (in the Minkowski hyperboloid model) is $\mathbb{H}^2 = {(x_1,x_2,x_3): x_3 > 0, x_3^2-x_1^2-x_2^2=1}$. This formulation uses three coordinates, while two are sufficient. The Minkowski hyperboloid model is a popular choice among computational geometers because distances and isometries are easy to compute in this model. The distance between two points $a = (x_1, x_2, x_3)$ and $a' = (x_1', x_2', x_3')$ is $\delta(a, a') = \mathrm{acosh}(x_3x_3' - x_1x_1' - x_2x_2')$. Importantly, the circumference and area of hyperbolic disks grow exponentially with their radius, providing sufficient space to represent tree-like and hierarchical structures with low distortion. Other common models of $\mathbb{H}^2$, obtained by projection into $\mathbb{R}^2$, include Beltrami-Klein disk model, Poincar\'e disk model, and native polar coordinates.
Although all of these models describe the same isometric abstract metric space, they differ in their numerical and visualization properties~\cite{numeric_pub}. In this paper, for simplicity, we focus on two-dimensional settings where points can be represented by radial and angular coordinates in $\mathbb{H}^2$, providing the geometric basis for the popularity--similarity decomposition introduced below.

\paragraph{Network geometry} In recent years, latent space models have gained attention due to their ability to produce synthetic networks with scale-free degree distributions, the small-world property, self-similarity, and a high clustering coefficient, among other properties~\cite{serrano2008self,krioukov2010hyperbolic,boguna2021network}.
One of the most widely used models is the $\mathbb{H}^2$ model~\cite{krioukov2010hyperbolic}. Each node $i$ is assigned polar coordinates $(r_i,\theta_i)$ in a hyperbolic disk of radius $R_{\mathbb{H}^2}$. Two nodes $i$ and $j$ are connected with probability
\begin{align}\label{eq:fermi_dirac}
    p_{ij} = \frac{1}{1 + \exp\left({\frac{x_{ij} - R_{\mathbb{H}^2}}{2T}}\right)},
\end{align}
where $T$ is the temperature parameter, which controls the level of clustering, and $x_{ij}$ is the hyperbolic distance between the two nodes. This distance is determined by
\begin{align}\label{eq:distance}
    \cosh x_{ij} = \cosh r_i \cosh r_j  - \sinh r_i \sinh r_j \cos \Delta\theta_{ij},
\end{align}
where $\Delta\theta_{ij} = \pi - |\pi - |\theta_i - \theta_j||$ denotes the angular distance between nodes $i$ and $j$.

The purely geometric $\mathbb{H}^2$ model is isomorphic to the $\mathbb{S}^1$ model~\cite{serrano2008self}. Under this correspondence, the radial coordinate of each node is mapped to a hidden degree $\kappa_i$, which represents the node's popularity or importance: $\kappa_i = \kappa_0 \exp((R_{\mathbb{H}^2} - r_i)/2)$ where $\kappa_0$ corresponds to the smallest hidden degree. This transformation allows the connection probability in Eq.~\eqref{eq:fermi_dirac} to be written as $p_{ij} = \frac{1}{1 + \chi^\beta}$ with $\chi = \frac{R\Delta\theta_{ij}}{\mu\kappa_i\kappa_j}$, where $\beta=1/T$ controls clustering and $\mu$ determines the average degree. The $\mathbb{H}^2$ and $\mathbb{S}^1$ formulations are therefore equivalent. The choice between them typically depends on whether a geometric representation or a hidden-variable representation is more convenient for the application under consideration.

\paragraph{Hyperbolic embedding} 
An embedding is the inclusion of one mathematical structure within another. 
Given a network $G=(V, E)$, where $V$ is the set of vertices and $E$ is the set of edges, its embedding into some geometry $\mathcal{M}$ is a map $m: V \rightarrow \mathcal{M}$. In particular, in \emph{hyperbolic} embeddings, $\mathcal{M}$ is hyperbolic space. \emph{Embedders} are algorithms used to solve that task. We can group them into four main types:
\begin{itemize}
    \item \textbf{Maximum-likelihood embedders (MLEs).} These methods infer node coordinates by maximizing the likelihood that the observed network was generated by a given hyperbolic network model. Representative methods include Mercator~\cite{garcia2019mercator,jankowski2023d} and  KVK~\cite{kitsak2020link} inspired by the early work of Boguñá et al.~\cite{boguna2010sustaining}, HyperMap~\cite{papadopoulos2015network} and HyperMapCN~\cite{papadopoulos2014network}, BFKL~\cite{blasius2016bfkl}, DHRG~\cite{celinskakopczynska2022dhrg}, and Anneal~\cite{celinska2024anneal}.
    \item \textbf{Structural or ordering-based embedders (SOEs).} These methods recover a hyperbolic layout using graph-theoretic properties, such as node similarity, local proximity, node ordering, and community structure. Representative methods include Coalescent~\cite{muscoloni2017coalescent}, CLOVE~\cite{balogh2025clove}, LPCS~\cite{wang2016lpcs}, and HMCS~\cite{wang2019fast}. 
    \item \textbf{Representation-learning embedders (RLEs).} These methods learn node representations by directly optimizing a task-specific loss function in hyperbolic space, typically using Riemannian gradient-based optimization. Representative methods include Poincar\'e embeddings~\cite{nickel2017poincare} and Lorentz embeddings~\cite{nickel2018lorentz}. 
    \item \textbf{Hybrid embedders (HEs).} These methods combine representation learning with model-based likelihood optimization. Typically, coordinates obtained using a representation-learning method are used to initialize a likelihood-based embedder, which subsequently refines the embedding according to a hyperbolic network model. Representative methods include Poincar\'e+DHRG and Lorentz+DHRG~\cite{celinska-kopczynska2026bridging}. RLEs have also been combined with SOEs in prior work~\cite{gerald2023hyperbolic}.
\end{itemize}
The preferred embedding approach varies across research communities. In network science, MLEs and SOEs are most commonly used. In machine learning, RLEs and SOEs are generally preferred, whereas algorithmic research tends to favor MLEs and hybrid methods.

Benchmarking in machine learning typically evaluates graph-learning methods on task-specific objectives, increasingly using synthetic networks to isolate structural effects under controlled conditions~\cite{morris2020tudataset,hu2020open,palowitch2022graphworld,maekawa2022beyond,aliakbarisani2026hypbench}. Synthetic benchmarks are also widely used in network science and have been applied to graph embedding~\cite{lancichinetti2008benchmark,zhang2021systematic}. Recent work has compared hyperbolic embedders across research communities~\cite{zhou_benchmark_2025, celinska-kopczynska2026bridging}. Here, we instead evaluate them on link prediction and topology reconstruction and study their adaptation to network incompleteness.

\section{Methods}
\begin{figure}[h]
    \centering
    \includegraphics[width=0.83\linewidth]{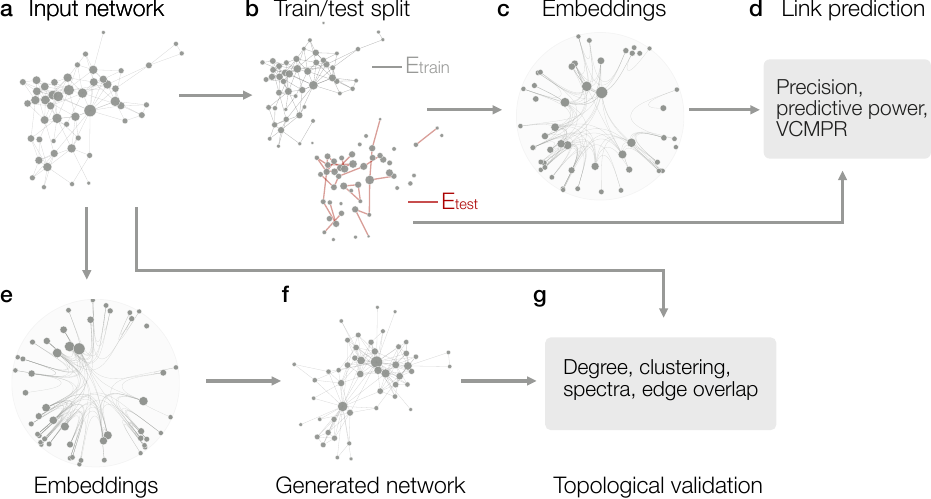}
    \caption{\textbf{Schematic view of the benchmarking pipeline.}  (\textbf{a}) We start from an input network $G=(V,E)$. (\textbf{b}) For the link prediction task, we randomly remove a fraction of edges, obtaining a training network $G_{\mathrm{train}}$ and a held-out test set $E_{\mathrm{test}}$. (\textbf{c}) We then embed $G_{\mathrm{train}}$ in hyperbolic space using different embedders. (\textbf{d}) Using the inferred node coordinates, candidate missing links are ranked by hyperbolic distance and evaluated against $E_{\mathrm{test}}$ using link-prediction measures. (\textbf{e}) For topology reconstruction, we embed the input network directly and (\textbf{f}) use the inferred coordinates and parameters $R$ and $T$ to generate synthetic networks. (\textbf{g}) Finally, we compare their local and global structural properties with those of the input network.}
    \label{fig:schematic}
\end{figure}

In this work, we use both synthetic and empirical networks to evaluate hyperbolic embedding methods on two tasks: link prediction and topology reconstruction. The evaluation pipeline is illustrated in Figure~\ref{fig:schematic}. We begin with an input network $G=(V,E)$. For link prediction, we construct a training network $G_{\mathrm{train}}$ by randomly removing a fraction $q$ of the edges. The removed edges form the held-out set used for evaluation. We then embed $G_{\mathrm{train}}$ using each hyperbolic embedding method, obtaining a set of coordinates for its nodes. Finally, we rank the candidate missing edges according to the hyperbolic distances between their endpoints and evaluate the resulting link-prediction performance. For topology reconstruction, we embed the full network $G$. Using the inferred node coordinates, we generate five networks (20 for real networks) and compare their topological properties with those of the original network. The following sections describe each step in detail.

\paragraph{Hyperbolic embedders} We selected 13 hyperbolic embedding methods from three different research communities. Our analysis is limited to methods specifically designed for unipartite networks, computationally feasible in practice, and supported by open-source implementations available by the end of 2025. When several methods represented successive refinements of the same underlying approach, we retained only the most recent version.
Following the classification we introduced in Section~2, we consider the following methods: Mercator~\cite{garcia2019mercator,jankowski2023d}, KVK~\cite{kitsak2020link}, BFKL~\cite{blasius2016bfkl}, DHRG~\cite{celinskakopczynska2022dhrg}, and Anneal~\cite{celinska2024anneal} as MLEs; Coalescent~\cite{muscoloni2017coalescent}, CLOVE~\cite{balogh2025clove}, LPCS~\cite{wang2016lpcs}, and HMCS~\cite{wang2019fast} as SOEs; Poincaré~\cite{nickel2017poincare} and Lorentz~\cite{nickel2018lorentz} as RLEs; and Poincaré+DHRG~\cite{celinska-kopczynska2026bridging} and Lorentz+DHRG~\cite{celinska-kopczynska2026bridging} as HEs. Some embedding methods, such as KVK, naturally incorporate missing links into their connection-probability model. In this work, we follow the same methodology as for KVK~\cite{kitsak2020link} and adapt Mercator, Anneal, and DHRG to account for network incompleteness (see Appendix~\ref{app:incomplete_conn_prob} for details). The remaining methods are not directly compatible with this type of adjustment. Appendix~\ref{apx:embedders} summarizes the hyperbolic embedders, and Appendix~\ref{apx:implementation} provides implementation details.


\paragraph{Task 1. Link prediction}
Because some embedding methods require a connected input graph, both embedding and evaluation are restricted to the giant connected component of $G_{\mathrm{train}}$. Each candidate pair $(i,j)\notin E_{\mathrm{train}}$ is scored according to its hyperbolic distance $d_{\mathbb{H}^2}(i,j)$, defined in Eq.~\eqref{eq:distance}. Candidate pairs are ranked by increasing distance, or equivalently, by decreasing score $s_{ij}=-d_{\mathbb{H}^2}(i,j)$. 
As a topology-based baseline, we also evaluate the Common Neighbors (CN) heuristic, which scores each candidate pair by the number of neighbors shared by its endpoints in $G_{\mathrm{train}}$.
We evaluate link-prediction performance using global precision, local precision, predictive power~\cite{cannistraci2013link}, and a vertex-centric metric~\cite{menand2024link}. More details are provided in Appendix~\ref{apx:lp}.

\paragraph{Task 2. Topology reconstruction}
For each embedding, we compute the connection-probability matrix $\mathbf{P}=(p_{ij})$ using Eq.~\eqref{eq:fermi_dirac}. 
Only a few method can infer $R_{\mathbb{H}^2}$. For consistency, however, we choose it such that the expected number of edges matches that of $G$, i.e., $\left|\sum_{i<j}p_{ij}-|E|\right|<\epsilon$.
For synthetic networks, we fix $T$ to its ground-truth value because Mercator is the only method considered that infers it directly. We then generate $G^\prime=(V,E^\prime)$ by sampling each node pair $(i,j)$ independently with probability $p_{ij}$. We then assess how well each embedding preserves the topology of the reference graph by comparing a collection of local and global structural observables measured on $G$ and on the corresponding synthetic realization $G^\prime$. The comparison measures are summarized in Appendix~\ref{apx:topology}. 
Hyperbolic coordinates alone do not define a generative model. Topology reconstruction requires coupling them with an explicit connection-probability function, such as the $\mathbb{S}^1/\mathbb{H}^2$ model used here, a capability not provided natively by many machine-learning embeddings.

\paragraph{Synthetic networks}
We generate synthetic networks from the $\mathbb{S}^1/\mathbb{H}^2$ model using the efficient generation procedure of Bl\"asius et al.~\cite{blasius2022efficient}. The benchmark spans network sizes $N\in\{500,1000,2000\}$, temperatures $T\in\{0.1,0.4,0.7\}$, average degrees $\langle k\rangle\in\{10,20\}$, and a power-law degree exponent $\gamma=2.5$. For each parameter combination, we generate 50 independent network realizations. 
The target average degree $\langle k\rangle$ is controlled through the average-degree parameter of the generator, denoted by $C$ in the implementation. In the native RHG notation, this role is played by the disk-radius offset $C$, with $R=2\log N+C$, which determines the expected average degree. For link prediction, after generating the complete synthetic graph, we remove a fraction $q\in\{0.1,0.5\}$ of edges uniformly at random to construct the training graph.

\paragraph{Real networks} We consider 11 datasets covering different domains. Since topology reconstruction requires temperature $T$, we infer it using Mercator. Table~\ref{tab:real_networks} in the Appendix summarizes their properties.

\section{Results}

\subsection{Effect of incompleteness on inferred network geometry}

Among the hyperbolic embedders considered, Mercator is the only one that infers both the node coordinates and the model parameters, including the temperature $T$, or equivalently, $\beta = 1/T$. Mercator infers $T$ by matching the input network’s clustering coefficient to that predicted by the model before angular coordinates are inferred; see Appendix A.3 of~\cite{garcia2019mercator} for more details.
As $T\to\infty$, the clustering coefficient approaches zero. Conversely, as $T \to 0$, Eq.~\eqref{eq:fermi_dirac} approaches a step function, such that nodes separated by less than a threshold distance are always connected.
However, because $T$ is inferred from the observed network, missing links may bias its estimated value. In this section, we derive the relationship between the inferred temperature $T$ and the fraction of missing links $q$.

Suppose that each edge is retained independently with probability $\rho=1-q$, where $q$ is the missing-link fraction. Denoting the connection probability in the fully observed network by $p_{ij}^{\mathrm{full}}$, the corresponding probability in the observed, incomplete network is $p_{ij}^{\mathrm{obs}}=\rho p_{ij}^{\mathrm{full}}$.
The thinned connection probability cannot be represented exactly by an unadjusted $\mathbb{S}^1$ model because its maximum connection probability is $\rho$ rather than one. Nevertheless, using a midpoint-matching approximation and assuming that the relevant pairwise geometric scales are approximately preserved after inference, we obtain the following implicit approximation for the effective inverse temperature $\beta'$:
\begin{align}\label{eq:beta_prime_main}
    \beta'
    \left[
    \log \rho
    +
    \log\left(
    \beta \sin\left(\frac{\pi}{\beta}\right)
    \right)
    -
    \log\left(
    \beta' \sin\left(\frac{\pi}{\beta'}\right)
    \right)
    \right]
    = \log\left(\frac{2-\rho}{\rho}\right).
\end{align}
The complete derivation and an empirical assessment of its assumptions are
provided in Appendix~\ref{apx:beta_vs_q}.

\begin{figure}[h]
    \centering
    \includegraphics[width=0.8\linewidth]{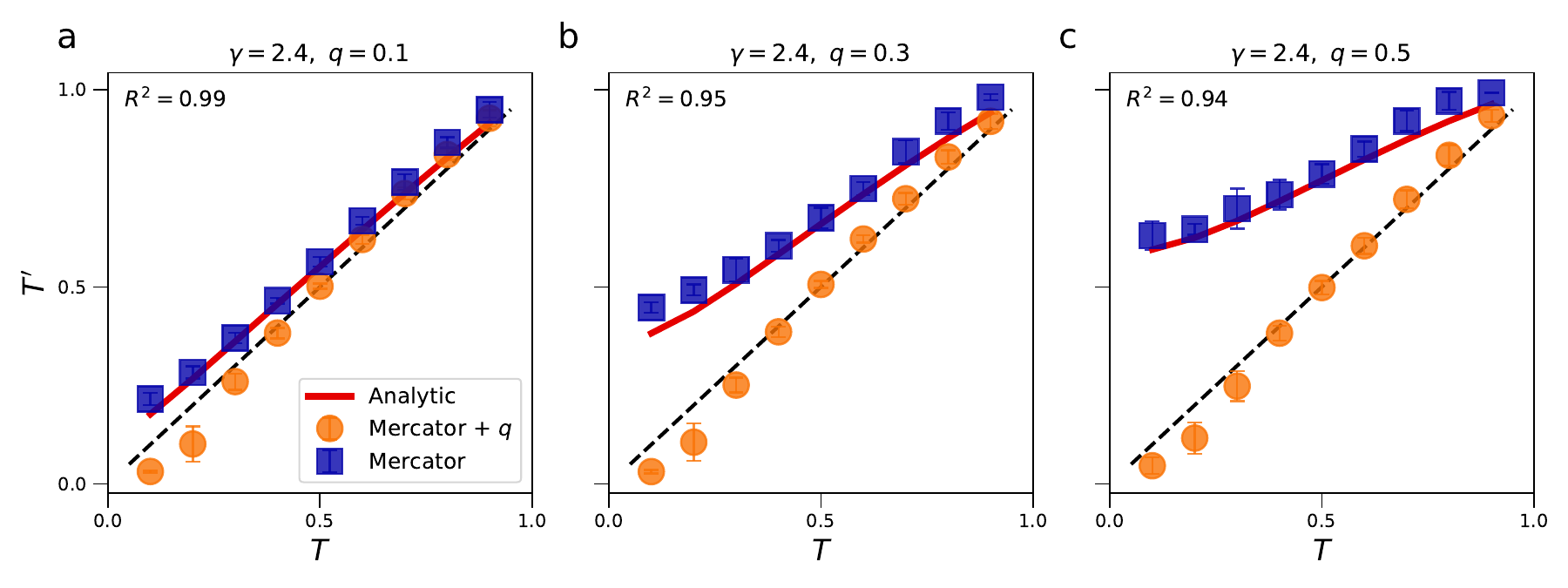}
    \caption{Comparison of input temperature $T=1/\beta$ with Mercator-inferred $T^\prime =1/\beta^\prime$. Red: Eq.~\eqref{eq:beta_prime_main}; blue: standard Mercator; orange: $q$-adjusted Mercator. $N=10^4, \langle k\rangle=10$; averaged over 10 realizations. The $R^2$ value in the upper-left of each panel is computed between the red analytic curve and the blue standard-Mercator squares across the plotted $T$ values.}
    \label{fig:beta_vs_beta_prime}
\end{figure}

Figure~\ref{fig:beta_vs_beta_prime} compares the analytical prediction with temperatures inferred by standard and incompleteness-aware Mercator. As the missing-link fraction increases, the standard Mercator infers a progressively larger effective temperature, whereas the version that explicitly accounts for $q$ approximately recovers the input temperature.
These results indicate that incompleteness affects the benchmark through two mechanisms: it removes topological information and can also distort the inferred geometric parameters. Consequently, a high inferred temperature in an observed network should not automatically be interpreted as evidence of intrinsically weak geometry when the network may be incomplete.

\subsection{Link prediction}
Following the pipeline shown in Figure~\ref{fig:schematic}a-d, we perform the link-prediction task on synthetic networks with varying parameters and on a set of real networks. 

\begin{figure}[h]
    \centering
    \includegraphics[width=0.9\linewidth]{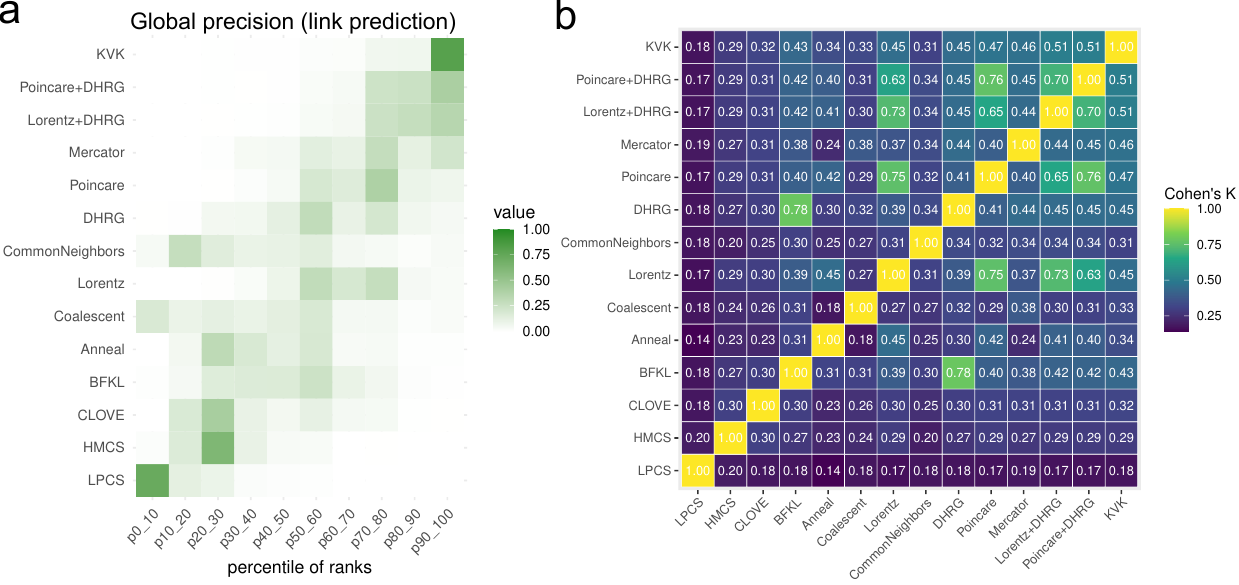}
    \caption{\textbf{Quality assessment of link prediction in synthetic networks}. (\textbf{a}) Aggregate rankings: darker colors indicate that the given embedder occurred more frequently in the given percentile of ranks (higher percentiles are better) over all synthetic graphs benchmarked. Embedders are ordered by the value in the last percentile. (\textbf{b}) Cohen's kappa coefficients for the agreement between pairs of embedders’ predictions. Higher values indicate higher prediction similarity.}
    \label{fig:lp_synthetic}
\end{figure}

\paragraph{Synthetic networks} 
First, we identify which hyperbolic embedding method performs consistently well across all network properties. Figure~\ref{fig:lp_synthetic}a presents the overall ranking of the hyperbolic embedders, averaged across all network parameter settings. KVK has the highest probability of being placed in the top percentile of the rankings, followed by the two hybrid methods, Poincaré+DHRG and Lorentz+DHRG. LPCS and HMCS rank lowest. In addition, Figure~\ref{fig:lp_synthetic_other} in the Appendix presents rankings based on local precision, predictive power, and VCMPR. These results generally lead to similar conclusions.
We also examine the relationship between link-prediction performance and network properties in greater detail. Overall, precision increases as the temperature decreases, that is, as the clustering coefficient increases, as the network becomes more incomplete, corresponding to larger values of $q$, and as the network becomes denser (see Figures~\ref{fig:lp_global_precision_appendix}--\ref{fig:lp_vcmpr_appendix} in the Appendix).

Second, we compare the similarity of the predictions produced by the hyperbolic embedders, examining whether they tend to predict the same links or different ones. In Figure~\ref{fig:lp_synthetic}b, we present the values of Cohen's kappa for pairs of embedders. Cohen's kappa is a statistical measure of inter-rater agreement between two raters classifying the same items into nominal categories. Unlike simple percentage agreement, Cohen’s kappa corrects for agreement expected by chance. In our case, agreement between the embedders is mostly fair to moderate (kappa values between 0.2 and 0.5), except for the high agreement between hybrid embedders and their base counterparts. This means different embedders exploit topological information differently, and some edges may be harder to predict than others. However, examining edge characteristics does not provide meaningful insights: the distributions of these characteristics between edges on which all embedders failed and those on which at least one embedder predicted correctly do not differ significantly.

\begin{figure}[h]
    \centering
    \includegraphics[width=0.8\linewidth]{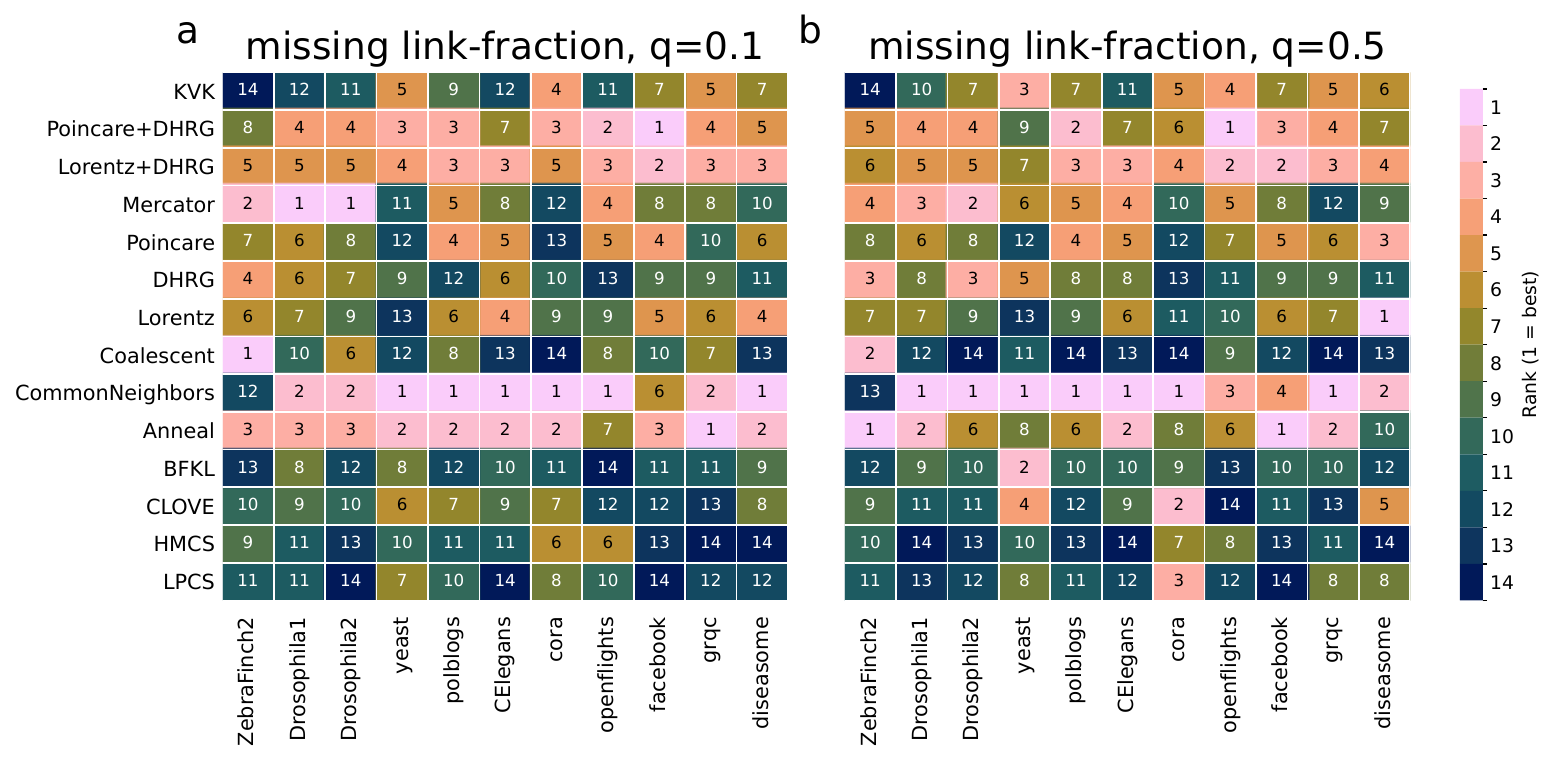}
    \caption{\textbf{Link prediction in real networks.} Ranking of the methods based on the median precision for each real network for missing link-fractions (\textbf{a}) $q=0.1$ and \textbf{(b)} $q=0.5$. Lower ranks are better. The order of embedders is set by their ranking in the link prediction task, as shown in Figure~\ref{fig:lp_synthetic}a.}
    \label{fig:lp_real}
\end{figure}

\paragraph{Real networks} Figure~\ref{fig:lp_real} presents the ranking of embedders by precision on real networks. The ranking remains relatively stable when performance is evaluated using predictive power (Figure~\ref{fig:lp_real_predictive_power}), local precision (Figure~\ref{fig:lp_real_local_precision}) or a vertex-centric metric (Figure~\ref{fig:lp_real_vertex}).
The top four positions are occupied by Common Neighbour, Anneal, Poincare+DHRG, and Lorentz+DHRG. This ordering differs slightly from that observed for synthetic networks, where KVK ranked first. One possible explanation is that real networks do not always conform closely to the assumed underlying hyperbolic model. Several of the analyzed networks have power-law degree-distribution exponents greater than 3, indicating more homogeneous degree distributions and a relative absence of large hubs. KVK takes this exponent as an input parameter but, by design, restricts it to values no greater than 3. This limitation may therefore contribute to its lower ranking on real networks. These structural differences may also partly explain why the CN heuristic performs comparatively well on real networks.
Conversely, Anneal, which ranked relatively low on synthetic networks, outperforms the other methods on several real networks, particularly on the connectome data that it was specifically designed to represent~\cite{celinska2024anneal}. 


\subsection{Topology reconstruction}
Following the pipeline shown in Figure~\ref{fig:schematic}a,e-g, we perform the topology reconstruction task on synthetic networks with varying parameters and on a set of real networks.

\begin{figure}[h]
    \centering
    \includegraphics[width=0.85\linewidth]{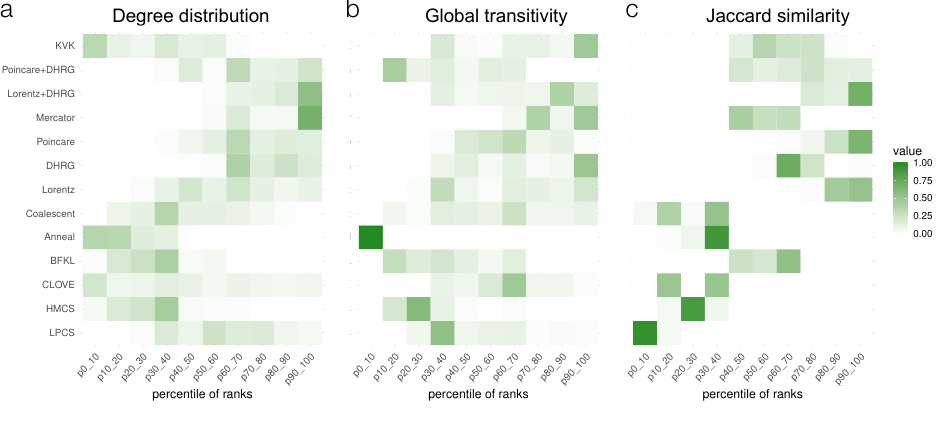}
    \caption{\textbf{Topology reconstruction in synthetic networks.}  Aggregated rankings for (\textbf{a}) Jensen-Shannon divergence between original and generated degree distributions, \textbf{(b)} absolute error in global transitivity between the original and generated networks, \textbf{(c)} Jaccard similarity between the original and generated edge sets. Darker colors indicate higher frequency in a given percentile of ranks. The order of embedders is set by their ranking in the link prediction task, as shown in Figure~\ref{fig:lp_synthetic}a.}
    \label{fig:tr_synthetic}
\end{figure}

\paragraph{Synthetic networks} Figure~\ref{fig:tr_synthetic} presents the rankings of the hyperbolic embedders according to their ability to reproduce the degree distribution, global transitivity, and Jaccard similarity between the edge sets of the reference and generated networks. Across all network parameter settings, Mercator and Lorentz+DHRG perform best in generating synthetic networks with degree distributions that closely match those of the reference networks. Similarly, Mercator, KVK, and DHRG reproduce global transitivity particularly well. Results for additional network properties are reported in Figure~\ref{fig:tr_synthetic_other}, while more detailed results, summarized by network size, temperature, and mean degree, are shown in Figures~\ref{fig:tr_degree_appendix}--\ref{fig:tr_jaccard_appendix}.

It is worth highlighting Figure~\ref{fig:tr_synthetic}c, which shows the overlap between the edge sets of the reference and generated networks. For example, Lorentz+DHRG tends to generate edges that are also present in the reference graph. By contrast, Mercator typically generates synthetic replicas that closely reproduce aggregate network properties but differ substantially from the reference graph at the level of individual edges. This distinction indicates that accurately reproducing global network statistics does not necessarily imply recovering the same underlying connectivity pattern.

\begin{figure}[h]
    \centering
    \includegraphics[width=0.9\linewidth]{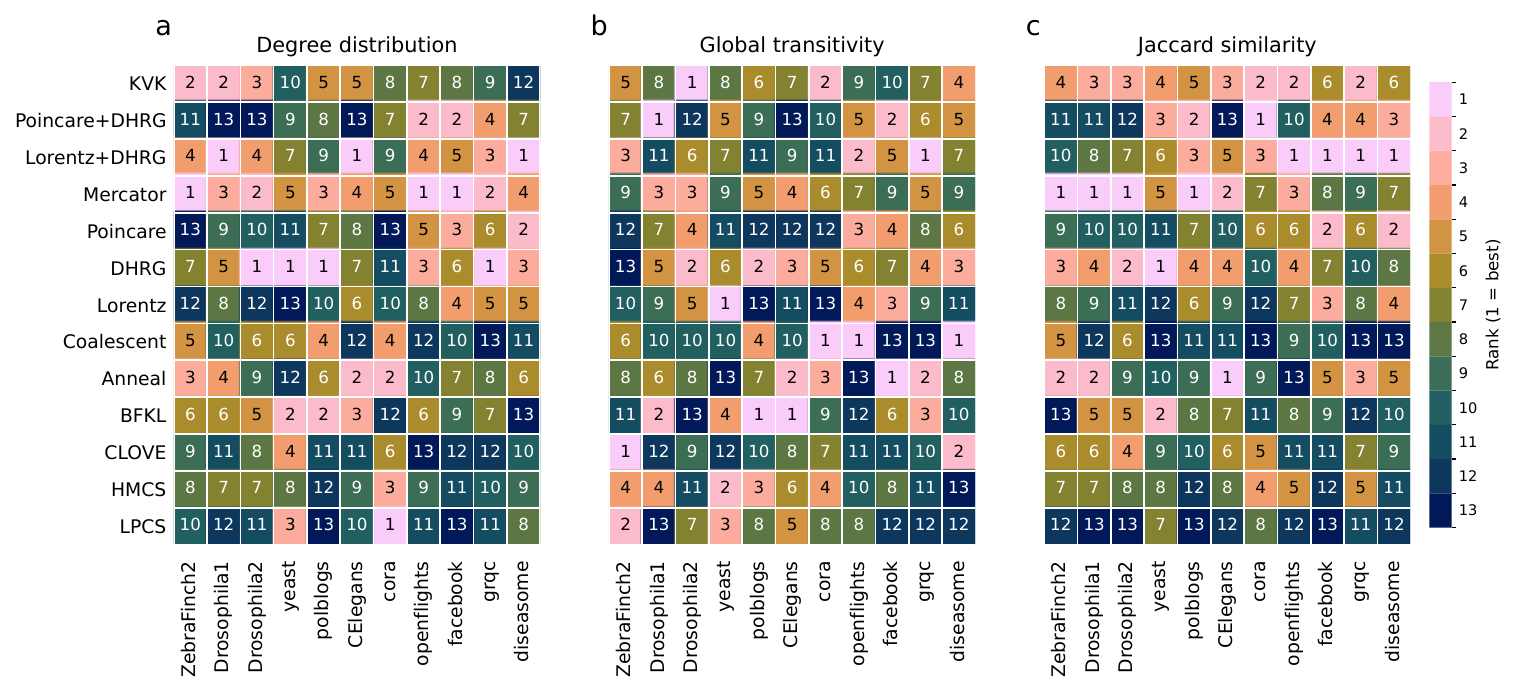}
    \caption{\textbf{Topology reconstruction in real networks.}  Rankings of the methods based on the (\textbf{a}) Jensen-Shannon divergence between the degree distribution of the original and generated graph, \textbf{(b)} absolute error in global transitivity, and \textbf{(c)} Jaccard similarity.
    For panels (a) and (b), rank 1 corresponds to the lowest error; for panel (c), rank 1 corresponds to the highest Jaccard similarity. The order of embedders is set by their ranking in the link prediction task, as shown in Figure~\ref{fig:lp_synthetic}a.}
    \label{fig:tr_real}
\end{figure}

\paragraph{Real networks} Figure~\ref{fig:tr_real} shows the performance of the embedders on the topology reconstruction task, and Figure~\ref{fig:real_networks_viz} visualizes representative original and generated networks. Mercator and DHRG generally reproduce the degree distribution well, although their ability to match global transitivity varies across real-world networks. Mercator and Lorentz+DHRG achieve the highest Jaccard similarity, indicating that their generated edge sets have the greatest overlap with the original edge sets among the evaluated embedders. The remaining three measures are reported in Figure~\ref{fig:tr_real_apx}. And Figure~\ref{fig:tr_real_raw_values} reports mean topology-reconstruction scores with sample standard deviations.
As in the link-prediction task, embedder rankings differ somewhat between synthetic and real networks. Determining whether these differences arise systematically from specific network properties will require evaluating a broader set of real networks spanning a wider range of topological properties.

\subsection{Trade-off between link prediction performance and quality of topology reconstruction}

We further investigate the relationship between performance in link prediction (LP) and topology reconstruction (TR) in synthetic networks. To aggregate rankings across multiple measures, we use a two-step procedure. First, for each network and task, we compute the standard Borda count over the task-specific measures, treating the measures as voters and assigning points to each embedder according to its rank. The resulting score captures an embedder’s overall performance across the measures associated with that task. Second, we treat the networks as voters and aggregate their Borda rankings using the Copeland rule. Under this rule, embedders are compared pairwise and receive 1 point for each win, 0.5 points for each tie, and 0 points for each loss.

\begin{wrapfigure}{r}{0.47\textwidth}
    \vspace{-2.0em}
    \centering
    \includegraphics[width=0.47\textwidth]{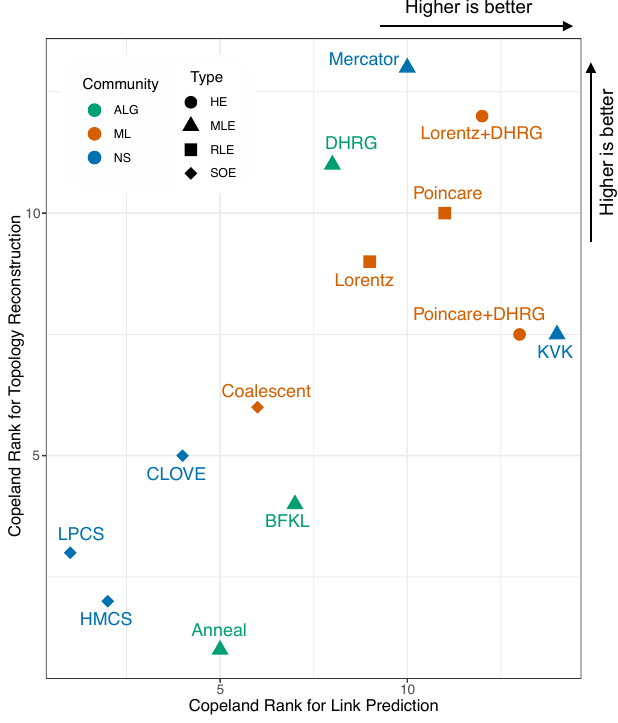}
    \caption{\textbf{Trade-off between link prediction and topology reconstruction in synthetic networks}. Ranks based on the Copeland method of aggregating ranks from multiple measures per task.} 
    \label{fig:tradeoff}
    \vspace{-3.5em}
\end{wrapfigure}

As shown in Figure~\ref{fig:tradeoff}, embedder rankings on the LP and TR tasks are positively correlated: embedders that perform well on one task generally also perform well on the other. However, this relationship is not fully predictive. KVK achieves the highest Copeland score for LP but ranks approximately sixth for TR, whereas Mercator achieves the highest score for TR but ranks only around fifth for LP. Thus, although performance across the two tasks is broadly aligned, the best-performing embedder depends on the specific task.

Performance also does not cluster by research community. Instead, it appears to be more closely associated with the underlying embedding paradigm. Maximum-likelihood-based embedders (MLEs), representation-learning-based embedders (RLEs), and hybrid embedders (HEs) generally outperform structural or ordering-based embedding methods (SOEs).

\section{Discussion and Limitations} 
Hyperbolic embedders are not universally superior for link prediction, and simple topology-based heuristics may outperform them on some networks. Their advantage is that they provide a reusable, unsupervised representation that can support multiple downstream tasks, including greedy routing and community detection.
All experiments use two-dimensional embeddings of unipartite networks. The $\mathbb{H}^{2}$ model can represent sparse, heterogeneous, and strongly clustered networks~\cite{krioukov2010hyperbolic}, while low-dimensional hyperbolic spaces can embed hierarchical structures with low distortion~\cite{nickel2017poincare}. We selected two dimensions because all evaluated methods support them. Several methods are formulated specifically in $\mathbb{H}^{2}$ or the equivalent $\mathbb{S}^{1}$ model. This common setting enables a consistent comparison but may disadvantage methods that could benefit from higher dimensions. Likewise, we considered only unipartite networks because bipartite, multiplex, and multilayer networks require specialized extensions available for only a subset of the methods. Our conclusions should therefore not be extrapolated directly to higher-dimensional or non-unipartite settings.
The benchmark also excludes computational cost and evaluates only link prediction and topology reconstruction. Moreover, reconstruction relies on the $\mathbb{S}^{1}/\mathbb{H}^{2}$ model, which does not explicitly represent community structure. Future work should compare runtime and memory requirements, higher-dimensional and non-unipartite methods, more expressive generative models such as nPSO~\cite{muscoloni2018nonuniform}, and additional tasks such as node classification and community detection.


\section{Conclusions}
We benchmarked 13 unsupervised hyperbolic embedders on link prediction and topology reconstruction. No method performed best across all networks and tasks. Performance was more closely tied to the embedding paradigm than to disciplinary origin: maximum-likelihood, representation-learning, and hybrid methods generally outperformed structural or ordering-based methods.
Performance on link prediction and topology reconstruction was positively related but not interchangeable. Some methods predicted missing edges well but reproduced network-level properties less accurately, whereas others provided better generative descriptions of the observed topology. Method selection should therefore account for both the downstream task and the input network's structure.
We also showed that random link removal can increase the temperature inferred by an unadjusted geometric model. Network incompleteness can therefore resemble weak latent geometry~\cite{van2022anomalous,van2024random} and should be considered when interpreting inferred geometric parameters.

\section*{Acknowledgments}
R.J. and M.K. acknowledge support from the Dutch Research Council (NWO) under grants OCENW.M20.244 and VI.C.242.10. We thank Eryk Kopczyński for helpful discussions. We thank Max Guichard for implementing a Mercator module for handling missing links.

\bibliographystyle{unsrtnat}
\bibliography{reference}

\newpage

\appendix
\section{Connection probability in the presence of missing links}\label{app:incomplete_conn_prob}

We follow the methodology introduced in~\cite{kitsak2020link}. Maximum-likelihood embedding methods estimate node coordinates by maximizing the likelihood
\begin{align}
\mathcal{L} = \prod_{i<j} p_{ij}^{a_{ij}} \left(1-p_{ij}\right)^{1-a_{ij}},
\end{align}
where $p_{ij}$ is the connection probability defined in Eq.~\eqref{eq:fermi_dirac}, and $a_{ij}$ denotes the corresponding entry of the observed adjacency matrix. This likelihood assumes that the network is fully observed.

Suppose instead that each existing link is missing independently with probability $q$. A link between nodes $i$ and $j$ is then observed only if it exists and is not removed. Consequently, the effective probability of observing a link is
\begin{align}
\tilde{p}_{ij}
=
(1-q)p_{ij}.
\end{align}
The likelihood of the observed network therefore becomes
\begin{align}
\mathcal{L}
=
\prod_{i<j}
\tilde{p}_{ij}^{a_{ij}}
\left(1-\tilde{p}_{ij}\right)^{1-a_{ij}}.
\end{align}
In this work, we adapt the Mercator, DHRG, and Anneal embedding methods to optimize this effective likelihood function.

\section{Hyperbolic embedders}\label{apx:embedders}

\begin{table}[h]
\caption{Comparison of hyperbolic embedders. Type: MLE -- Maximum-likelihood embedder, SOE -- Structural or ordering-based embedder, RLE -- Representation-learning embedder, HE -- Hybrid embedder. Community: NS -- Network Science, ALG -- Algorithms, ML -- Machine Learning. Hyperparameters: $\gamma$ -- exponent of the power-law degree distribution, $T =1/\beta$ -- temperature, $q$ -- fraction of missing links.}
\label{tab:hyp_embedders}
\centering
\small
\setlength{\tabcolsep}{2.5pt}
\renewcommand{\arraystretch}{1}
\begin{tabular}{@{}l c c c c@{}}
\toprule
Method & Type & Community & \makecell[c]{Explicitly models\\incompleteness?} & Hyperparameters \\
\midrule
KVK~\cite{kitsak2020link} & MLE  & NS & \yes  & $\gamma, T, q$ \\
Mercator~\cite{garcia2019mercator} & MLE  & NS & \yes (adapted)  & $q$ \\
CLOVE~\cite{balogh2025clove} &  SOE    & NS & \no  & community detection algorithm  \\
LPCS~\cite{wang2016lpcs}   &  SOE   & NS & \no   & $\gamma$, community detection algorithm \\
DHRG~\cite{celinskakopczynska2022dhrg} & MLE & ALG & \yes (adapted) & tessellation  \\
BFKL~\cite{blasius2016bfkl}  &   MLE   & ALG & \no  & $\gamma,T$ \\
Anneal~\cite{celinska2024anneal}  &  MLE  & ALG & \yes (adapted)  & tessellation, simulated annealing steps \\
Poincar\'e~\cite{nickel2017poincare} & RLE & ML & \no & learning rate $\eta$, dimension $D$, epochs  \\
Poincar\'e+DHRG~\cite{celinska-kopczynska2026bridging} & HE & ML/ALG & \no   & learning rate $\eta$, dimension $D$, epochs  \\
Lorentz~\cite{nickel2018lorentz} & RLE & ML & \no & learning rate $\eta$, dimension $D$, epochs \\
Lorentz+DHRG~\cite{celinska-kopczynska2026bridging}  & HE &  ML/ALG & \no  & learning rate $\eta$, dimension $D$, epochs\\
Coalescent~\cite{muscoloni2017coalescent} & SOE & ML/NS & \no  & \makecell{pre-weighting rule, angular adjustment,\\nonlinear dimension reduction} \\
HMCS~\cite{wang2019fast} & SOE & NS & \no   & number of nested hierarchies \\
\bottomrule
\end{tabular}
\end{table}

Table~\ref{tab:hyp_embedders} summarizes the hyperbolic graph embedders used in this work. Network incompleteness can affect every method we consider here. We do not assume that all methods are immune to its effects. The relevant distinction is whether the sampling process can be incorporated into a method’s optimization objective without fundamentally altering the method.

We adapted all embedding methods for which such a correction was possible and retained the original formulations of the remaining methods. Mercator, Anneal, and DHRG are maximum-likelihood methods whose objective functions depend directly on the connection probability $p_{ij}$. If each existing link is observed independently with probability $q$, then the probability of observing a link is $q p_{ij}$. Network incompleteness can therefore be incorporated by replacing $p_{ij}$ with $q p_{ij}$ in the likelihood, following the approach used in KVK.

The remaining methods optimize objectives for which there is no equivalent way to incorporate $q$ without changing the methods themselves. We therefore evaluated these baselines using their original formulations. For example, the Poincaré embedder uses the connection probability in its link-prediction score but its embedding objective does not maximize the corresponding likelihood of the observed network. Consequently, the same correction cannot be applied directly.

We explicitly highlight that in the column \textit{Explicitly models incompleteness}, which describes whether a method was originally designed for network incompleteness, was adapted in this work, or could not be corrected to account for it.

\section{Implementations used}\label{apx:implementation}

In general, we follow the framework published in the supplementary material of~\cite{celinska-kopczynska2026bridging}. Here, we summarize our setup.


We have downloaded the embedders from the following repositories and used the following settings:

\begin{itemize}
\item Poincar\'e and Lorentz: \url{https://github.com/facebookresearch/poincare-embeddings} (last commit on Sep 16, 2021),
Attribution-NonCommercial 4.0 International

We use the hyperparameters
{\tt -epochs 1500 -negs 50 -burnin 20 -dampening 0.75 -ndproc 4 -eval\_each 100 -fresh -sparse -burnin\_multiplier 0.01 -neg\_multiplier 0.1 -lr\_type constant -lr 1 -train\_threads 1 -dampening 1.0 -batchsize 50 -gpu 0}
from the example {\tt train-nouns.sh} from the repository, except that we requested using the GPU ({\tt -train\_threads 1 -gpu 0}).
We also add the hyperparameters specifying a method ({\tt -manifold poincare -dim 2}).
For Lorentz 2D, Poincar\'e 3D, Poincar\'e 5D, Lorentz 3D, we replace {\tt -lr 1} with {\tt -lr 0.5 -no-maxnorm} (this setting comes from the suggestion about Lorentz embeddings in {\tt train-nouns.sh}).

\item BFKL: \url{https://bitbucket.org/HaiZhung/hyperbolic-embedder/overview} (last commit on Sep 8, 2016), no license given

This method estimates the hyperparameters in {\tt estimateHyperbolicParameters} method. We do not modify the original settings.
For real networks, the temperature ($T$) parameter for embedding is set based on Mercator's inferred value,
the parameter $\alpha$ is estimated based on fitting the power law, and the radius ($R$) is computed using a formula.

\item DHRG: \texttt{rogueviz/dhrg} subdirectory in the RogueViz engine

This method is parameterized by the tessellation used; we use the bitruncated order-3 heptagonal tiling.
It does not create embeddings from scratch, but rather improves them using local search; we allow up to 110 iterations of local search. Local search computes the log-likelihood using the logistic function.


\item Mercator: \url{https://github.com/networkgeometry/mercator} (last commit Jun 21, 2022), GPL v3

We use the full version of Mercator that includes maximum-likelihood optimization. We do post-processing of the inferred values of the radial positions.
The parameter $\beta$ can be provided, but we use the default behavior, in which $\beta$ is inferred to reproduce the average local clustering coefficient of the original edgelist.



\item Simulated annealing: \texttt{rogueviz/sag} subdirectory in the RogueViz engine.

This method is parameterized by the tessellation used; we use the bitruncated order-3 heptagonal tiling for 2D embeddings,
and the subdivided(2) \{4,3,5\} honeycomb for 3D embeddings (the {\tt g711} and {\tt g435b2} settings from the original paper).
As in the original paper, we set the parameter controlling the number of tiles to $M = 20000$. The number of iterations of
simulated annealing is $N_S = 10000 |V|$. As in the original paper, we run the embedder twice; the first pass is to obtain
good initial values of the $R$ and $T$ parameters.

\item Coalescent: \url{https://github.com/biomedical-cybernetics/coalescent_embedding} (last commit Jul 8, 2019)

We use the hyperparameters and settings from {\tt RUN\_EXAMPLE.m}. Specifically, 2D embeddings use RA1-LE-EA. 3D embeddings use RA1-ISO.
We run the code in Octave (the free alternative of MATLAB), which has no access to {\tt graphallshortestpaths} function; we
solve this issue by computing the table of shortest paths with the C++ implementation from~\cite{celinska-kopczynska2026bridging}.

\item KVK: \url{https://bitbucket.org/dk-lab/2020_code_hyperlink/src/master/} (last commit Jun 11, 2011)

This embedder has two parameters: the power-law exponent $\gamma$ and the temperature $T$. For simulated networks
we use the actual temperature (providing more information to the embedder), whereas for real-world networks we use the temperature inferred by Mercator.
As explained in the paper, $\gamma=2\alpha+1$; we take $\alpha$ estimated by BFKL.

\item LPCS: the source code is included with the paper at \url{https://www.sciencedirect.com/science/article/pii/S0378437116000182}

The source code is in MatLab. As it is very slow, we apply the corrections and use C++ implementation from~\cite{celinska-kopczynska2026bridging}.

\item HMCS: We were also unable to find an official implementation of this embedder, so we use the reimplementation from~\cite{celinska-kopczynska2026bridging} to include the changes (obtaining the hierarchy by calling FMO algorithm repeatedly rather than just once; Community Closeness instead
of Community Intimacy; angular size of a community based on the sum of degrees rather than the number of vertices). This embedder has
one hyperparameter: the number of nested hierarchies to use. The authors suggest 2, or more for larger networks; we use 5 levels.

\item CLOVE: \url{https://github.com/samu32ELTE/hypCLOVE} (last commit Oct 16, 2025)

We use the default values of all settings and hyperparameters: $\gamma$ to fit the degree distribution,
degree fitting sample size of 100, automatically detected dendrogram, Leiden community detection method,
exponential coarsening, the number 1 of anchor communities, Christofides algorithm for solving the Travelling Salesman Problem,
{\tt degree\_greedy} node arrangement, community sector sizing based on the number of nodes in the community, and PSO radial coordinates assigned.

\end{itemize}

For reproducibility, we also control the PRNG seed.

We use the following hardware: 

[1] Intel\textregistered\ Core\texttrademark\ i7-9700K CPU @ 3.60GHz, NVIDIA GeForce GTX 1060 6GB/PCIe/SSE2, 96 GB RAM (we used zram for the embedders
which did not fit in RAM)

[2] 11th Gen Intel\textregistered\ Core\texttrademark\ i7-11850H @ 2.50GHz, OpenGL renderer string: NVIDIA RTX A3000 Laptop GPU/PCIe/SSE2

Software: Arch Linux, g++ versions 12.2.1 to 15.2.1 (DHRG, BFKL, KVK, Anneal, Mercator), Python 3.6 (Poincar\'e, Lorentz, ltiling, Mercator), Octave 10.3 (Coalescent, LPCS),
R 4.5.2 (LPCS, creation of graphs).

\section{Details of link prediction}\label{apx:lp}
Given an input network $G=(V, E)$, we randomly hold out a fraction $q$ of the edges for testing. This partitions the edge set into a training set $E_{\mathrm{train}}$ and a test set $E_{\mathrm{test}}$, yielding
\begin{align}
G_{\mathrm{train}}=(V,E_{\mathrm{train}}) \qquad\text{and}\qquad G_{\mathrm{test}}=(V,E_{\mathrm{test}}),
\end{align}
where $E_{\mathrm{train}}\cup E_{\mathrm{test}}=E$. After embedding $G_{\mathrm{train}}$, we rank all candidate edges, that is, node pairs not connected in $G_{\mathrm{train}}$, in ascending order of their hyperbolic distance. We then evaluate the top-ranked predictions against the held-out edge set $E_{\mathrm{test}}$.

\paragraph{Global precision} 
Let $L_m$ be the set of the top-$m$ ranked candidate edges, with $m=|E_{\mathrm{test}}|$. We report Precision@$m$,
\begin{equation}
\mathrm{Precision@}m \;=\; \frac{|L_m \cap E_{\mathrm{test}}|}{m},
\end{equation}
i.e., the fraction of held-out edges recovered among the top $m$ predictions.

\paragraph{Local precision} We also evaluate precision at a smaller prediction budget $b$, focusing only on the highest-ranked candidate edges:
\begin{equation}
\mathrm{Precision@}b \;=\; \frac{|L_b \cap E_{\mathrm{test}}|}{b},
\end{equation}
with $b<m$. Whereas $\mathrm{Precision@}m$ measures recovery at the full test-set scale, $\mathrm{Precision@}b$ emphasizes the quality of the very top predictions. Small values of $b$ therefore probe how well a method prioritizes the most likely missing edges. To summarize performance across multiple prediction budgets, we compute the averaged local precision over a set of budgets. Larger averaged local precision indicates that high precision is maintained consistently across a wider range of budgets.

\paragraph{Predictive power}
Following~\cite{cannistraci2013link}, the predictive power is defined as
\begin{align}
    PP = 10 \log_{10} \frac{\mathrm{Precision}@m}{\mathrm{Precision@m_{rand}}},
\end{align}
where $\mathrm{Precision}@m$ is the global precision and $\mathrm{Precision@m_{rand}}$ is the random precision. Predictive power is expressed in decibels, where larger values indicate better performance.

\paragraph{Vertex-centric metric} For each node $u$, we rank all candidate neighbors $v$ with $(u,v)\notin E_{\mathrm{train}}$ by increasing hyperbolic distance and keep the top-$k$ list $L_k(u)$, where $k=10$. Let
\begin{equation}
R(u)=\{v:(u,v)\in E_{\mathrm{test}}\}, \qquad d_{\mathrm{test}}(u)=|R(u)|,
\end{equation}
and define
\begin{equation}
t_u(k)=|L_k(u)\cap R(u)|.
\end{equation}
Then the vertex-centric maximum precision/recall metric (VCMPR@$k$)~\cite{menand2024link} is
\begin{equation}
\mathrm{VCMPR@}k(u)=\frac{t_u(k)}{\min\{k,d_{\mathrm{test}}(u)\}}.
\end{equation}
We report the average over nodes.

\section{Details of topology reconstruction}\label{apx:topology}
We use six measures to compare the topology of the input network $G=(V,E)$ with that of the generated network $G^\prime=(V,E^\prime)$:
\begin{itemize}
    \item \textbf{Degree distribution $P(k)$.} The degree distribution describes the probability that a node has degree $k$. We compute the Jensen--Shannon divergence between the degree distributions of $G$ and $G^\prime$. The divergence ranges from 0 to 1, where lower values indicate more similar distributions.
    \item \textbf{Clustering spectrum $c(k)$.} The clustering spectrum gives the average clustering coefficient of nodes with degree $k$. We compute the root-mean-square error (RMSE) between the clustering spectra of $G$ and $G^\prime$. Lower values indicate more similar spectra.
    \item \textbf{Average neighbor-degree spectrum $k_{\mathrm{nn}}(k)$.} The average neighbor-degree spectrum gives the mean degree of the neighbors of nodes with degree \(k\). It characterizes degree correlations and provides information about the network's assortative or disassortative structure. We compute the RMSE between the spectra of $G$ and $G^\prime$. Lower values indicate more similar spectra.
    \item \textbf{Spectral gap.} Let $\lambda_1(G)$ and $\lambda_2(G)$ denote the two largest eigenvalues of the adjacency matrix of $G$, ordered such that $\lambda_1(G) \geq \lambda_2(G)$. The spectral gap is defined as $\Delta\lambda(G)=\lambda_1(G)-\lambda_2(G)$. We compute the absolute difference between the spectral gaps of the two networks $\left|\Delta\lambda(G)-\Delta\lambda(G^\prime)\right|$.
    \item \textbf{Global transitivity.} Global transitivity measures the fraction of connected triples that form closed triangles. It is defined as $T(G)=\frac{3\,N_{\triangle}(G)}{N_{\wedge}(G)}$, where $N_{\triangle}(G)$ is the number of triangles and $N_{\wedge}(G)$ is the number of connected triples in $G$. We compute the absolute difference $\left|T(G)-T(G^\prime)\right|$. Lower values indicate more similar levels of global clustering.
    \item \textbf{Jaccard similarity.} We compute the Jaccard similarity between the edge sets $E$ and $E^\prime$: $J(E,E^\prime) =\frac{|E\cap E^\prime|}{|E\cup E^\prime|}$. The original and reconstructed graphs share the same vertex set, with a fixed one-to-one correspondence between node identifiers. Thus, the similarity is computed directly between their edge sets and is independent of the order in which the nodes are stored. We do not maximize the similarity over vertex permutations, as this would constitute an NP-hard graph-matching problem and would measure similarity between unlabeled graph structures rather than recovery of edges between the correct identified node pairs. Higher values indicate greater edge recovery, and $J(E,E^\prime)=1$ means that the two networks contain exactly the same edges between the same node pairs.
\end{itemize}

\newpage 
\section{Relationship between incompleteness and inverse temperature}\label{apx:beta_vs_q}

We derive how random link removal affects the effective inverse temperature
inferred when incompleteness is not explicitly modeled. For analytical
convenience, we use the $\mathbb{S}^1$ model, which is isomorphic to the
$\mathbb{H}^2$ model.

Let $p_{ij}^{\mathrm{full}}$ denote the connection probability in the fully
observed network:
\begin{align}
    p_{ij}^{\mathrm{full}}
    =
    \frac{1}{
        1+
        \left(
            \frac{R\Delta\theta_{ij}^{\mathrm{full}}}
                 {\mu_{\mathrm{full}}(\beta)
                  \kappa_i^{\mathrm{full}}
                  \kappa_j^{\mathrm{full}}}
        \right)^\beta
    },
    \label{eq:p_full}
\end{align}
where $\Delta\theta_{ij}^{\mathrm{full}}$ is the angular distance,
$\kappa_i^{\mathrm{full}}$ and $\kappa_j^{\mathrm{full}}$ are hidden degrees,
$R=N/(2\pi)$, and $\beta=1/T$ is the inverse temperature. The parameter
$\mu_{\mathrm{full}}(\beta)$ fixes the expected average degree
$\langle k\rangle_{\mathrm{full}}$:
\begin{align}
    \mu_{\mathrm{full}}(\beta)
    =
    \frac{\beta}
         {2\pi\langle k\rangle_{\mathrm{full}}}
    \sin\left(\frac{\pi}{\beta}\right).
    \label{eq:mu_full}
\end{align}
This expression applies in the regime $\beta>1$, equivalently $T<1$.

Suppose that each existing edge is retained independently with probability
$\rho=1-q$, where $q$ is the missing-link fraction. The connection
probability in the observed, thinned network is then
\begin{align}
    p_{ij}^{\mathrm{obs}}
    =
    \rho p_{ij}^{\mathrm{full}},
    \label{eq:p_obs}
\end{align}
and its expected average degree is
\begin{align}
    \langle k\rangle_{\mathrm{obs}}
    =
    \rho\langle k\rangle_{\mathrm{full}}.
    \label{eq:k_obs}
\end{align}
We ask whether the thinned network can be approximated by a standard
$\mathbb{S}^1$ model that does not explicitly account for missing links. Let
$\beta'=1/T'$ denote the effective inverse temperature obtained by fitting
such an unadjusted model to the observed network. Its fitted connection
probability is
\begin{align}
    p_{ij}^{\mathrm{fit}}
    =
    \frac{1}{
        1+
        \left(
            \frac{\xi_{ij}^{\mathrm{fit}}}
                 {\mu_{\mathrm{fit}}(\beta')}
        \right)^{\beta'}
    },
    \label{eq:p_fit}
\end{align}
where
\begin{align}
    \mu_{\mathrm{fit}}(\beta')
    &=
    \frac{\beta'}
         {2\pi\langle k\rangle_{\mathrm{obs}}}
    \sin\left(\frac{\pi}{\beta'}\right)
    \nonumber\\
    &=
    \frac{\beta'}
         {2\pi\rho\langle k\rangle_{\mathrm{full}}}
    \sin\left(\frac{\pi}{\beta'}\right).
    \label{eq:mu_fit}
\end{align}
For the fully observed and fitted geometries, respectively, define
\begin{align}
    \xi_{ij}^{\mathrm{full}}
    &\coloneqq
    \frac{R\Delta\theta_{ij}^{\mathrm{full}}}
         {\kappa_i^{\mathrm{full}}\kappa_j^{\mathrm{full}}},
    \label{eq:xi_full}
    \\
    \xi_{ij}^{\mathrm{fit}}
    &\coloneqq
    \frac{R\Delta\theta_{ij}^{\mathrm{fit}}}
         {\kappa_i^{\mathrm{fit}}\kappa_j^{\mathrm{fit}}}.
    \label{eq:xi_fit}
\end{align}
Equation~\eqref{eq:p_full} can therefore be written as
\begin{align}
    p_{ij}^{\mathrm{full}}
    =
    \frac{1}{
        1+
        \left(
            \frac{\xi_{ij}^{\mathrm{full}}}
                 {\mu_{\mathrm{full}}(\beta)}
        \right)^\beta
    }.
    \label{eq:p_full_xi}
\end{align}
The thinned connection law cannot be represented exactly by the unadjusted
fitted model when $\rho<1$. In particular,
\begin{align}
    p_{ij}^{\mathrm{obs}}\to\rho
    \qquad\text{as}\qquad
    \Delta\theta_{ij}^{\mathrm{full}}\to0,
\end{align}
whereas
\begin{align}
    p_{ij}^{\mathrm{fit}}\to1
    \qquad\text{as}\qquad
    \Delta\theta_{ij}^{\mathrm{fit}}\to0.
\end{align}
We therefore seek an effective value $\beta'$ that approximates the thinned
connection law rather than reproducing it exactly.

\begin{figure}[h]
    \centering
    \includegraphics[width=0.45\linewidth]{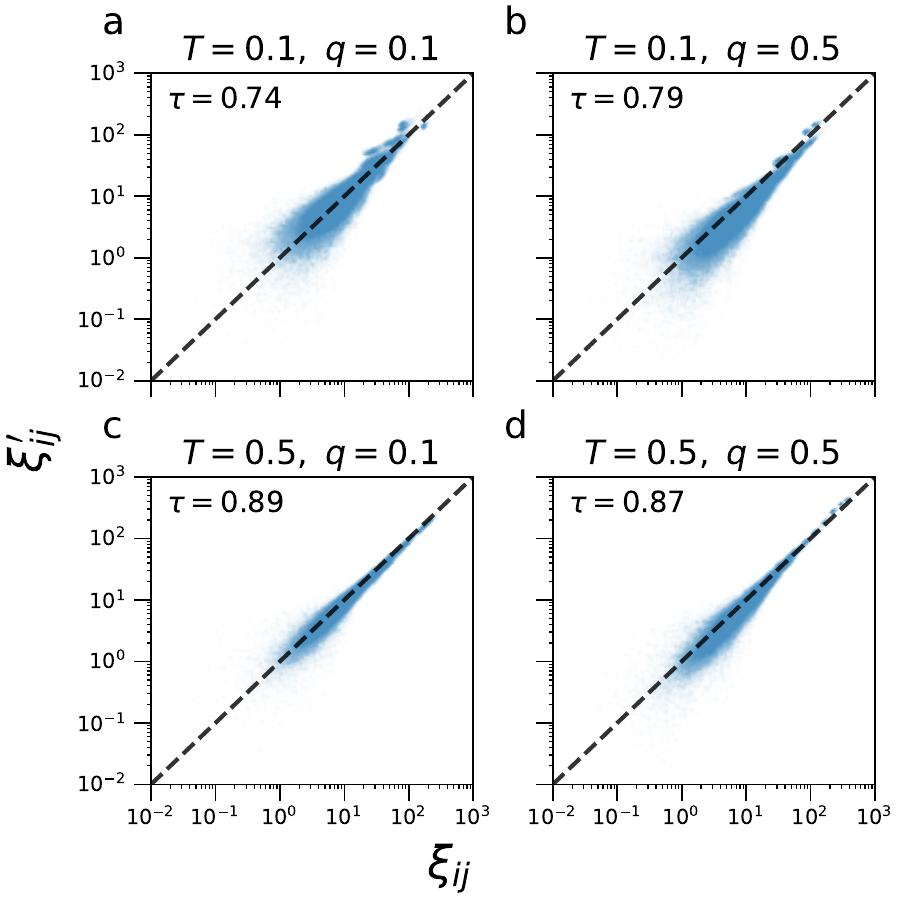}
    \caption{Comparison between $\xi_{ij}$ and $\xi_{ij}^\prime$ computed from the coordinates inferred by Mercator. Network parameters: $N=1000, \gamma=2.5$, (\textbf{a}) $T=0.1, q=0.1$, (\textbf{b}) $T=0.1, q=0.5$, (\textbf{c}) $T=0.5, q=0.1$, (\textbf{d}) $T=0.5, q=0.5$. In the top-left corner of each panel, we report the Kendall’s $\tau$ coefficient.}
    \label{fig:xi}
\end{figure}

Figure~\ref{fig:xi} shows that $\xi_{ij}^{\mathrm{full}}$ and $\xi_{ij}^{\mathrm{fit}}$ are strongly
correlated. Motivated by this observation, we assume that the relevant
pairwise geometric scales are approximately preserved near the transition
region of the connection probability:
\begin{align}
    \xi_{ij}^{\mathrm{fit}}
    \simeq
    \xi_{ij}^{\mathrm{full}}.
    \label{eq:xi_approx}
\end{align}

We estimate $\beta'$ by matching the observed and fitted connection
probabilities at the midpoint of the fully observed connection law. This
midpoint is
\begin{align}
    \xi_\star
    =
    \mu_{\mathrm{full}}(\beta),
\end{align}
at which
\begin{align}
    p^{\mathrm{full}}(\xi_\star)
    =
    \frac{1}{2},
    \qquad
    p^{\mathrm{obs}}(\xi_\star)
    =
    \frac{\rho}{2}.
    \label{eq:observed_midpoint}
\end{align}

Under the approximation in Eq.~\eqref{eq:xi_approx}, we impose
\begin{align}
    p^{\mathrm{fit}}(\xi_\star)
    =
    \frac{\rho}{2}.
\end{align}
Substituting Eq.~\eqref{eq:p_fit} gives
\begin{align}
    \frac{1}{
        1+
        \left(
            \frac{\mu_{\mathrm{full}}(\beta)}
                 {\mu_{\mathrm{fit}}(\beta')}
        \right)^{\beta'}
    }
    =
    \frac{\rho}{2},
\end{align}
and hence
\begin{align}
    \left(
        \frac{\mu_{\mathrm{full}}(\beta)}
             {\mu_{\mathrm{fit}}(\beta')}
    \right)^{\beta'}
    =
    \frac{2-\rho}{\rho}.
    \label{eq:midpoint_ratio}
\end{align}

Substituting Eqs.~\eqref{eq:mu_full} and \eqref{eq:mu_fit} into
Eq.~\eqref{eq:midpoint_ratio} yields
\begin{align}
    \left[
        \rho
        \frac{
            \beta\sin\left(\frac{\pi}{\beta}\right)
        }{
            \beta'\sin\left(\frac{\pi}{\beta'}\right)
        }
    \right]^{\beta'}
    =
    \frac{2-\rho}{\rho}.
    \label{eq:beta_prime_power}
\end{align}
Equivalently,
\begin{align}
    \boxed{
    \beta'
    \left[
        \log\rho
        +
        \log\left(
            \beta\sin\left(\frac{\pi}{\beta}\right)
        \right)
        -
        \log\left(
            \beta'\sin\left(\frac{\pi}{\beta'}\right)
        \right)
    \right]
    =
    \log\left(\frac{2-\rho}{\rho}\right)
    }.
    \label{eq:beta_prime}
\end{align}
Equation~\eqref{eq:beta_prime} can be solved numerically for $\beta'>1$. The corresponding effective temperature is $T'=1/\beta'$. In the fully observed limit $\rho=1$, the equation recovers $\beta'=\beta$ and hence $T'=T$.

\begin{figure}[h]
    \centering
    \includegraphics[width=0.85\linewidth]{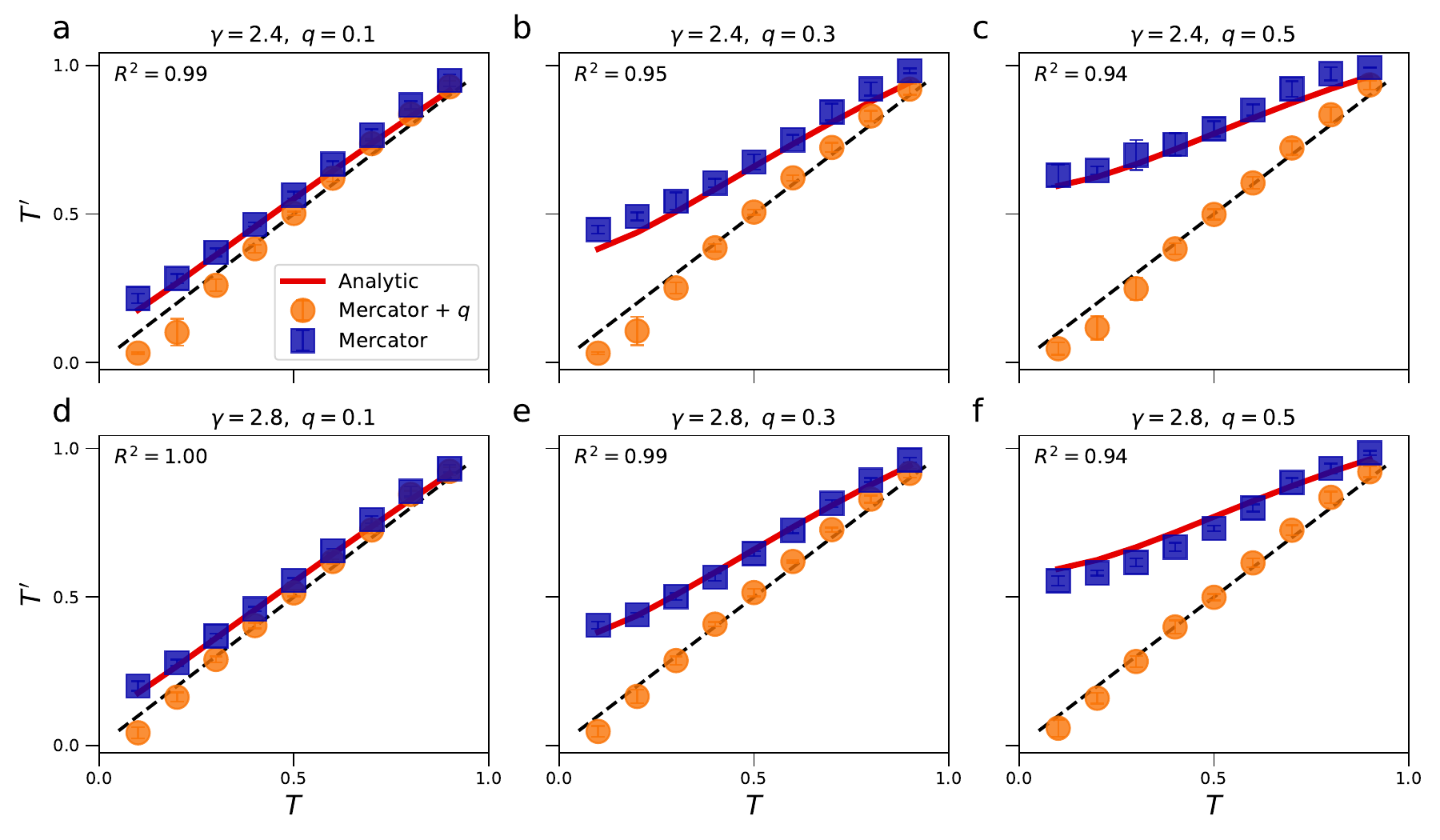}
    \caption{Comparison between the input temperature $T$ and the temperature $T^\prime$ inferred by Mercator. The red curve shows the analytical prediction from Eq.~\eqref{eq:beta_prime}; blue squares correspond to standard Mercator, without the $q$-adjustment; and orange circles correspond to the adjusted version of Mercator that accounts for $q$. Network parameters are $N=10000$ and $\langle k\rangle=10$. Results are averaged over 10 network realizations. The $R^2$ value in the upper-left of each panel is computed between the red analytic curve and the blue standard-Mercator squares across the plotted $T$ values.}
    \label{fig:analytics_large}
\end{figure}

Figure~\ref{fig:analytics_large} compares the midpoint approximation with the numerical results. The approximation closely follows the effective temperature $T'$ inferred by standard Mercator, which increases as the missing-link fraction grows. By contrast, the version of Mercator that explicitly accounts for $q$ approximately recovers the input temperature
$T$ of the fully observed network.

\newpage 
\section{Additional results for the synthetic networks}\label{apx:synthetic_networks}

\subsection{Link prediction results}

\begin{figure}[h]
    \centering
    \includegraphics[width=0.95\linewidth]{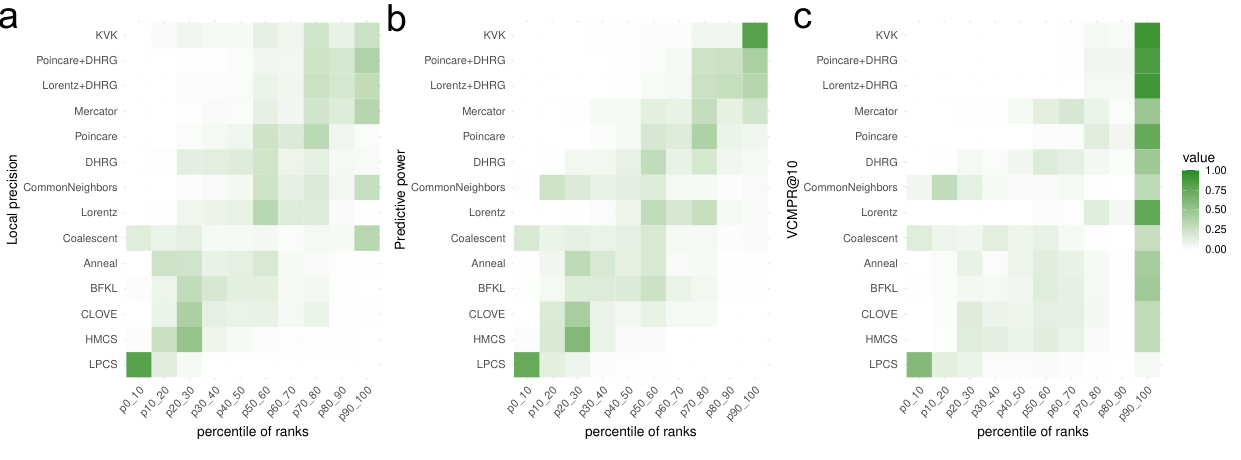} 
    \caption{Rankings of embedders for link prediction based on three measures: (\textbf{a}) local precision, (\textbf{b}) predictive power, and (\textbf{c}) VCMPR. Darker colors indicate that the given embedder occurred more frequently in the given percentile of ranks (higher percentiles are better) over all synthetic graphs benchmarked.}
    \label{fig:lp_synthetic_other}
\end{figure}

\begin{figure}[h]
    \centering
    \includegraphics[width=\linewidth]{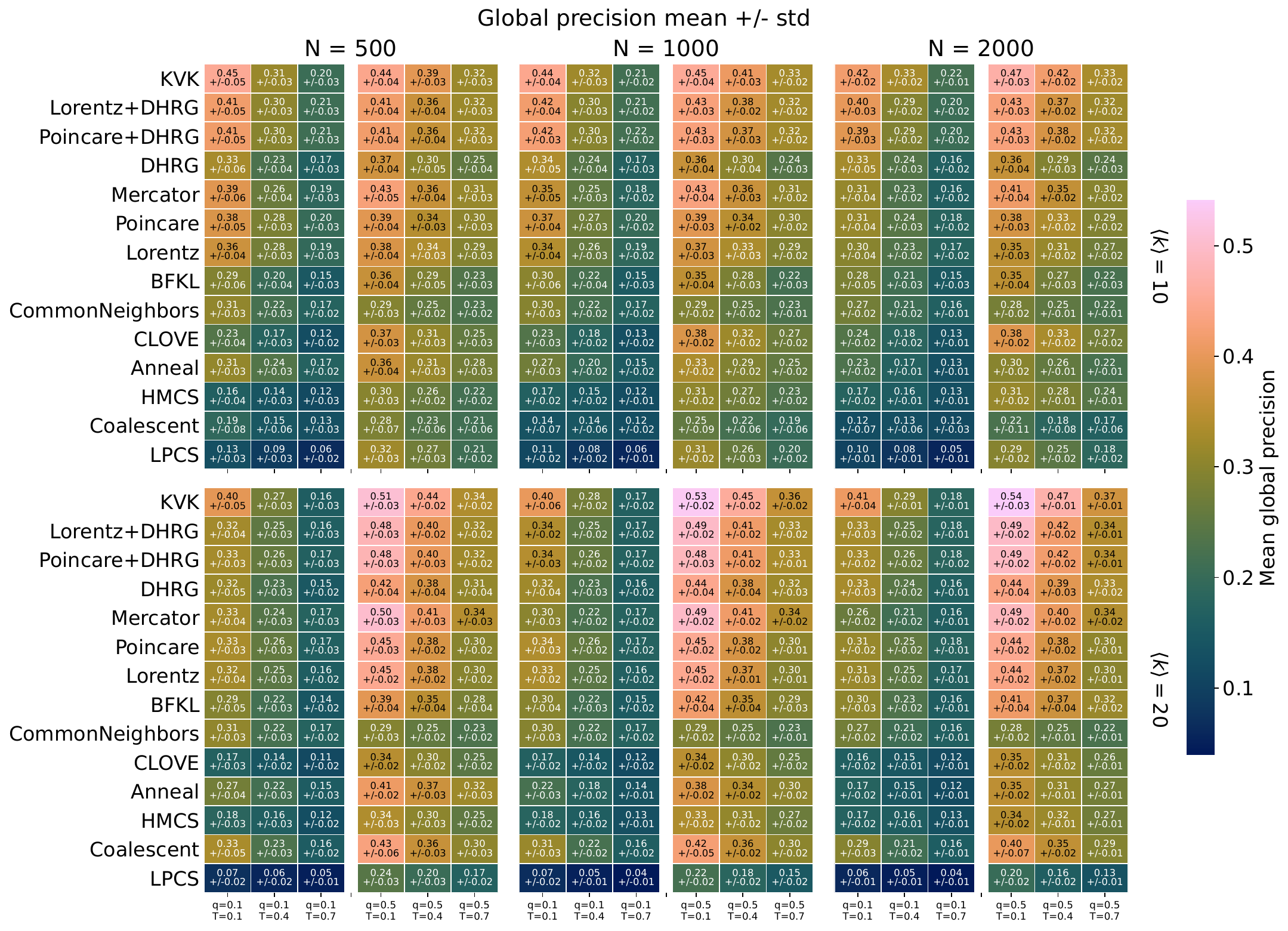} 
    \caption{Global precision in synthetic networks. The top panel corresponds to a network generated with $\langle k \rangle = 10$, and the bottom panel corresponds to a network generated with $\langle k \rangle = 20$. Each heatmap shows the performance of an embedding method at a given temperature $T$ and missing-link fraction $q$. Results are computed over 50 realizations, with the mean and standard deviation reported in each cell.}
    \label{fig:lp_global_precision_appendix}
\end{figure}

\newpage

\begin{figure}[h]
    \centering
    \includegraphics[width=0.88\linewidth]{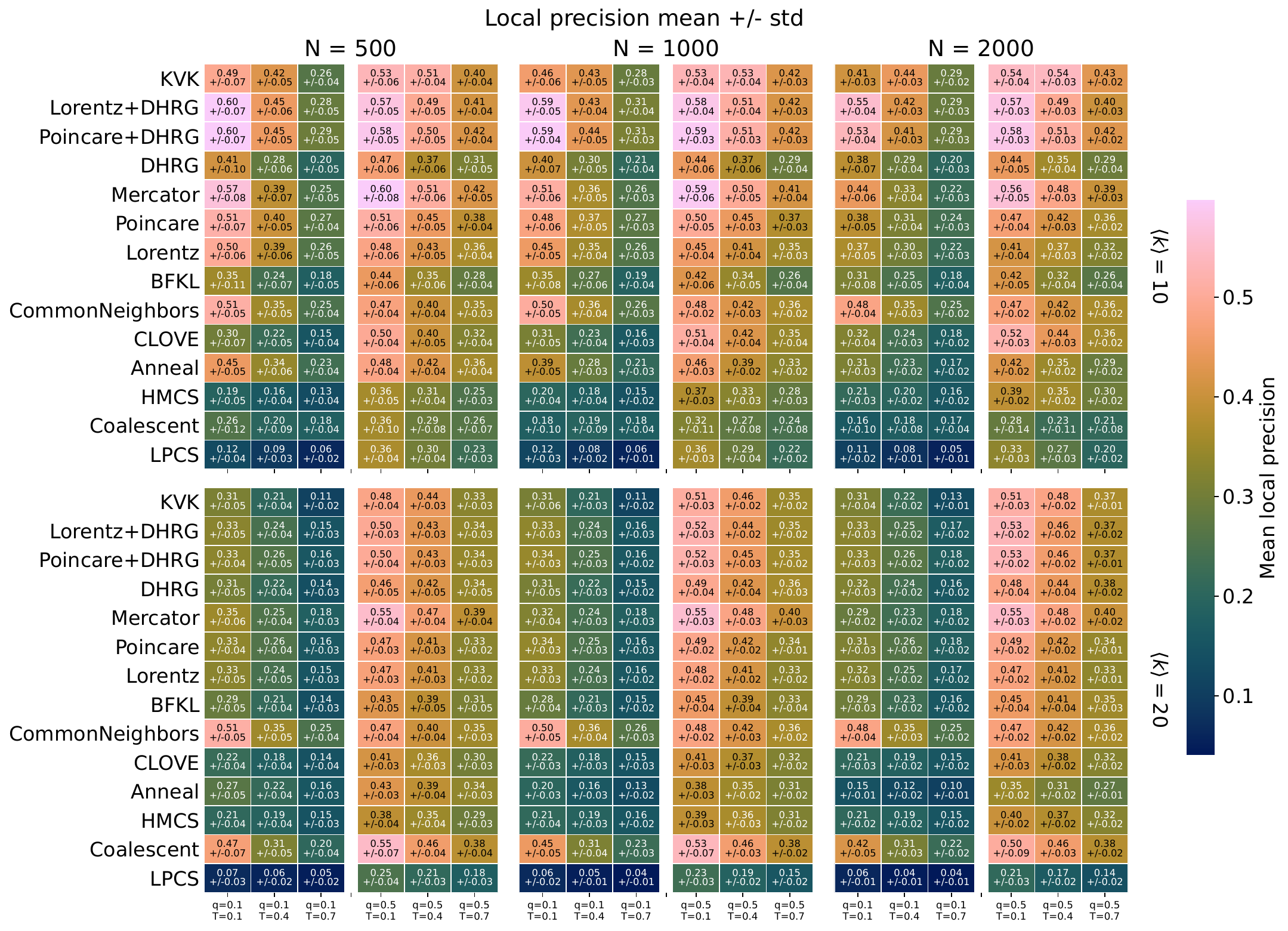} 
    \caption{Local precision in synthetic networks. See caption of Figure~\ref{fig:lp_global_precision_appendix} for more details.}
    \label{fig:lp_local_precision_appendix}
\end{figure}

\begin{figure}[h]
    \centering
    \includegraphics[width=0.88\linewidth]{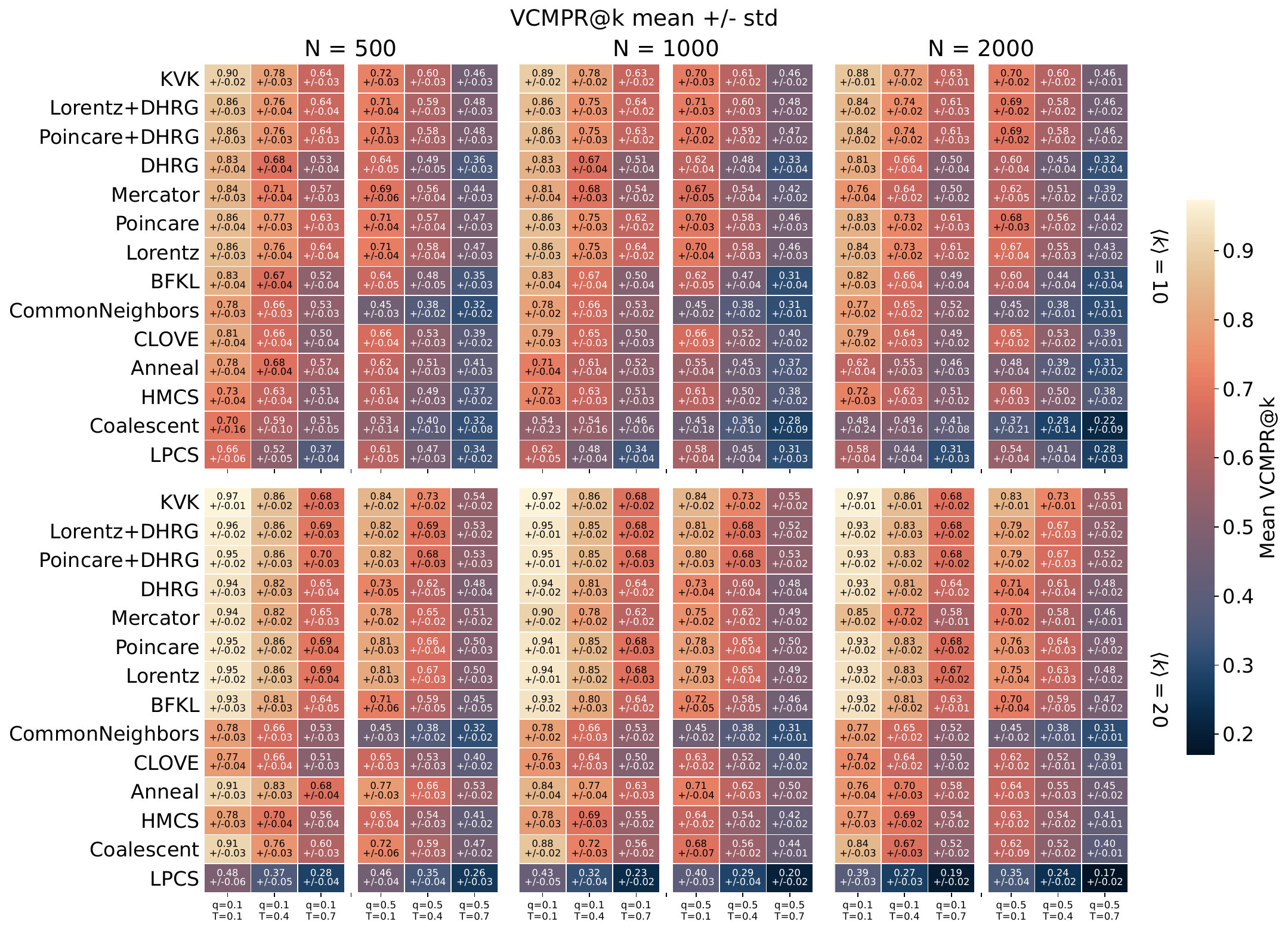} 
    \caption{VCMPR@$k$ in synthetic networks. See caption of Figure~\ref{fig:lp_global_precision_appendix} for more details.}
    \label{fig:lp_vcmpr_appendix}
\end{figure}

\newpage
\subsection{Topology reconstruction results}

\begin{figure}[h]
    \centering
    \includegraphics[width=0.32\linewidth]{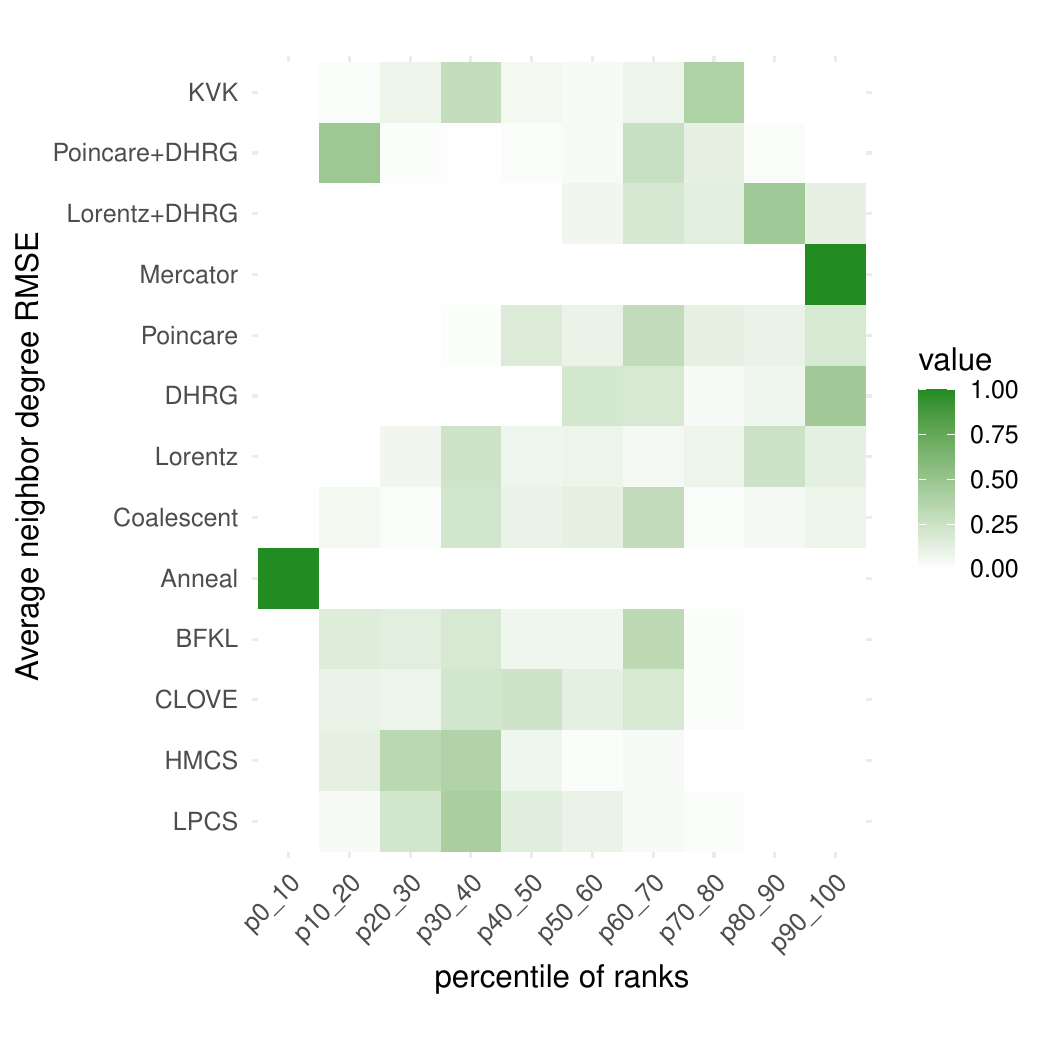}
    \includegraphics[width=0.32\linewidth]{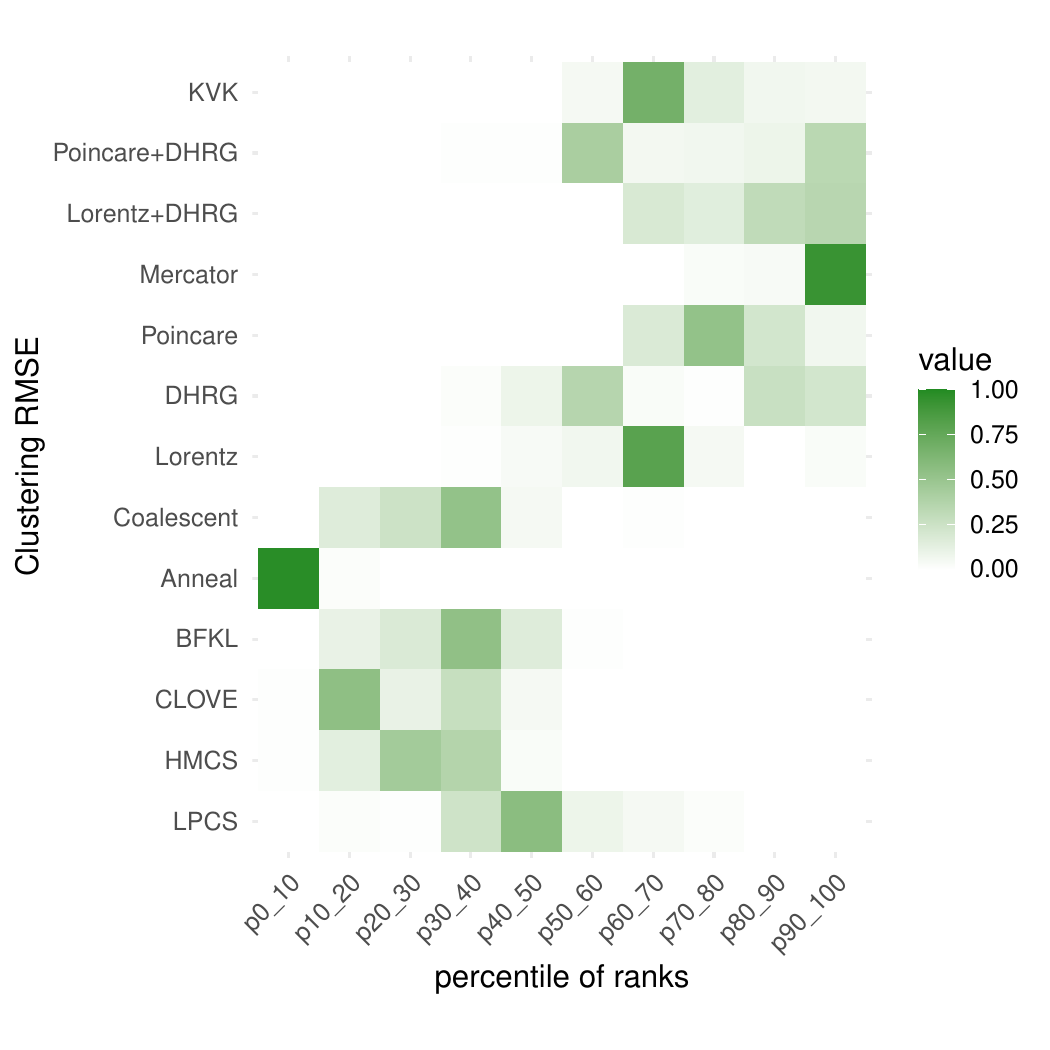}
    \includegraphics[width=0.32\linewidth]{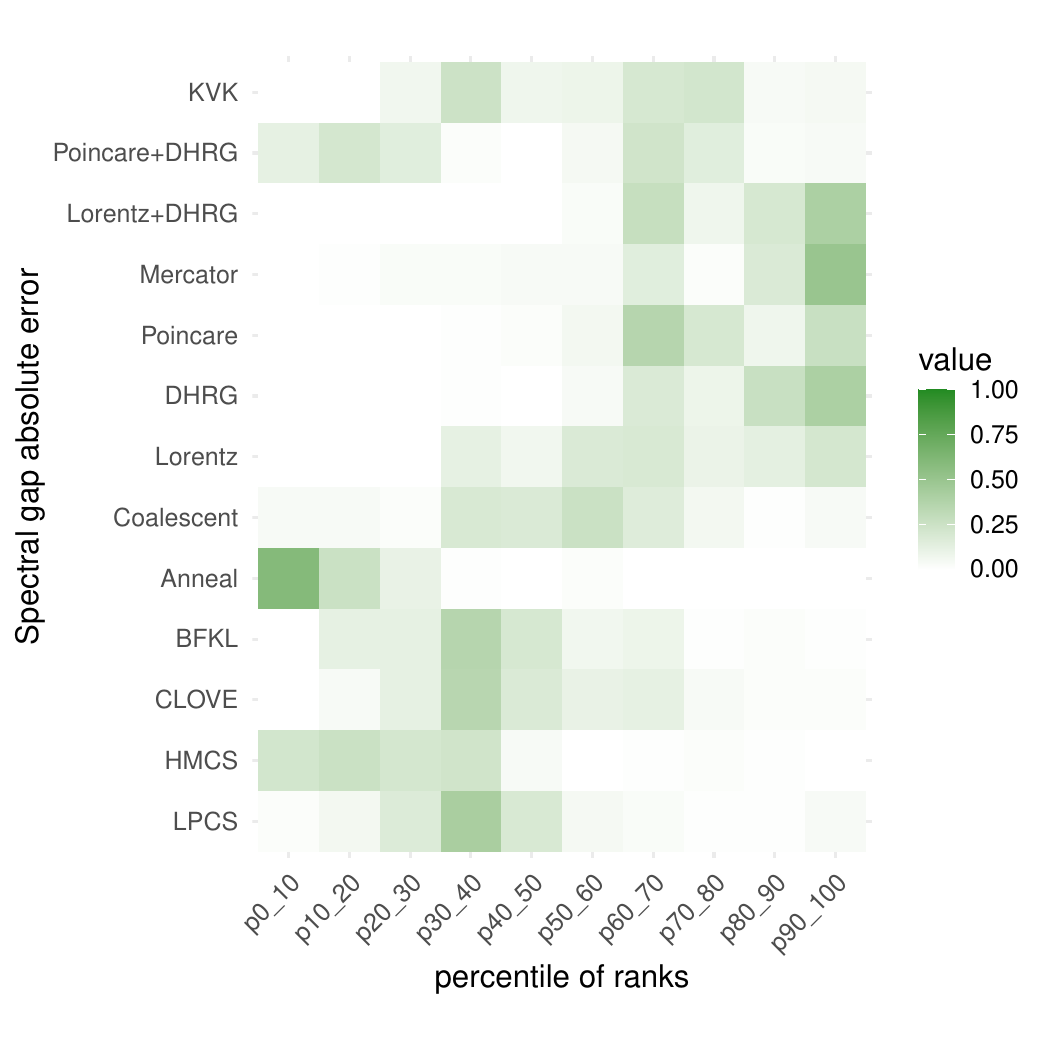}
    \caption{Additional topology-reconstruction rankings in synthetic networks. Aggregated rankings for (\textbf{a}) RMSE of the average neighbor-degree spectrum, (\textbf{b}) RMSE of the clustering spectrum, and (\textbf{c}) absolute error in the spectral gap. Darker colors indicate a higher frequency in a given percentile of ranks; higher percentiles are better.}
    \label{fig:tr_synthetic_other}
\end{figure}

\begin{figure}[h]
    \centering
    \includegraphics[width=0.7\linewidth]{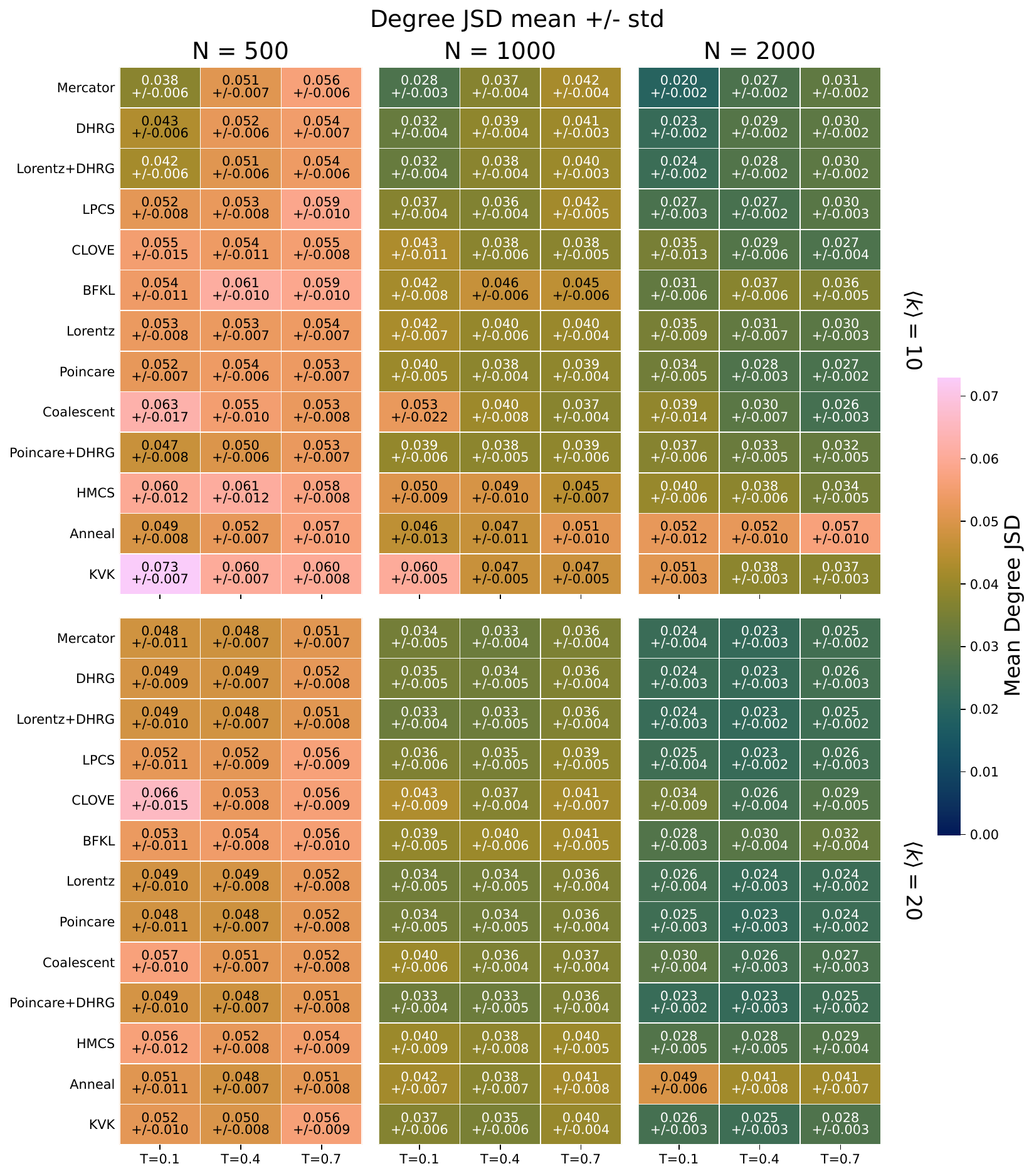} 
    \caption{Jensen-Shannon divergence between the original degree distribution and that of the generated synthetic networks at different model temperatures. Lower values indicate closer agreement between the two distributions.}
    \label{fig:tr_degree_appendix}
\end{figure}

\newpage
\begin{figure}[h]
    \centering
    \includegraphics[width=0.9\linewidth]{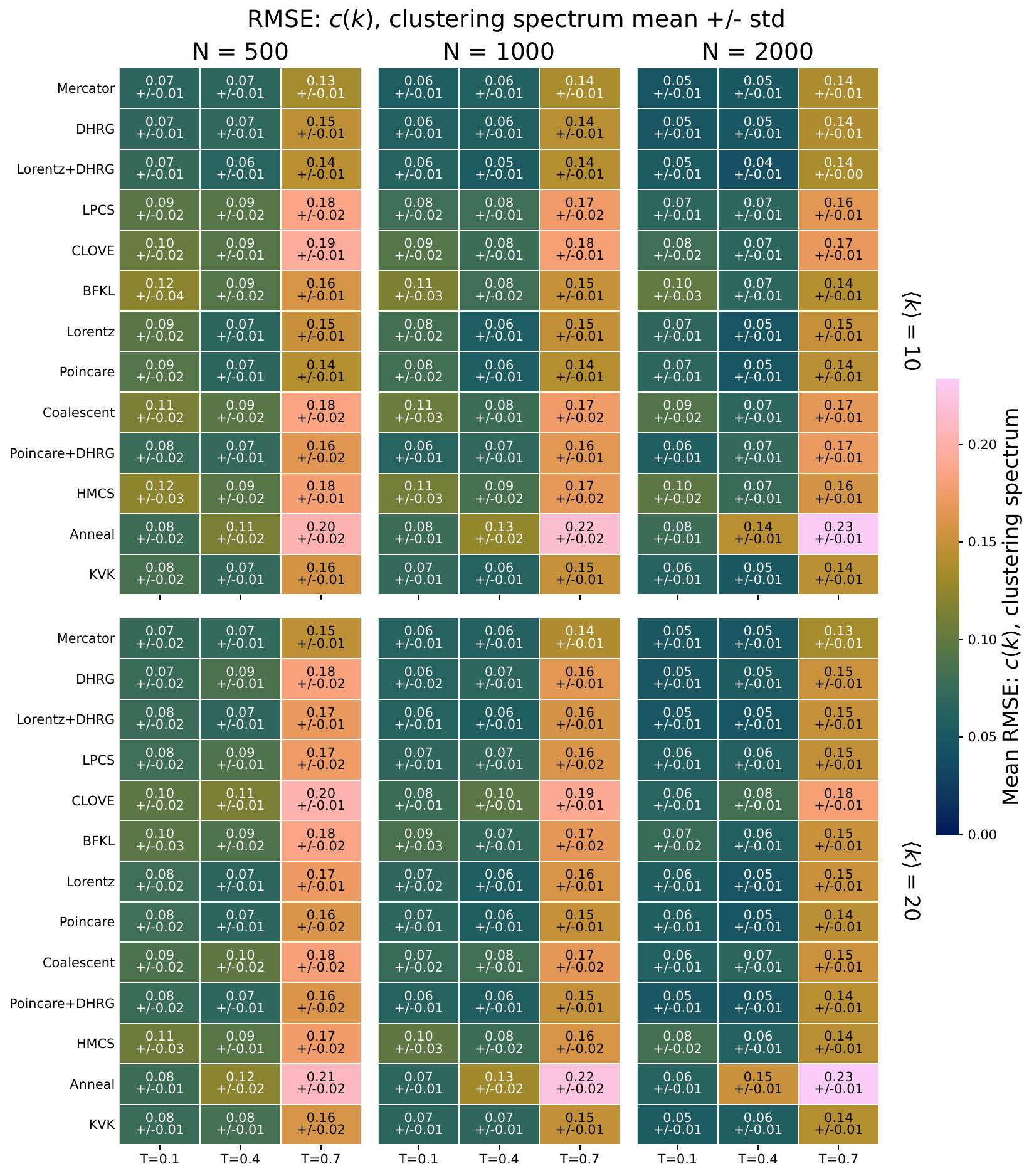} 
    \caption{Root-mean-square error (RMSE) between the clustering spectrum of the original graph and of the generated synthetic networks at different model temperatures. Lower values indicate closer agreement.}
    \label{fig:tr_ck_appendix}
\end{figure}

\newpage
\begin{figure}[h]
    \centering
    \includegraphics[width=0.9\linewidth]{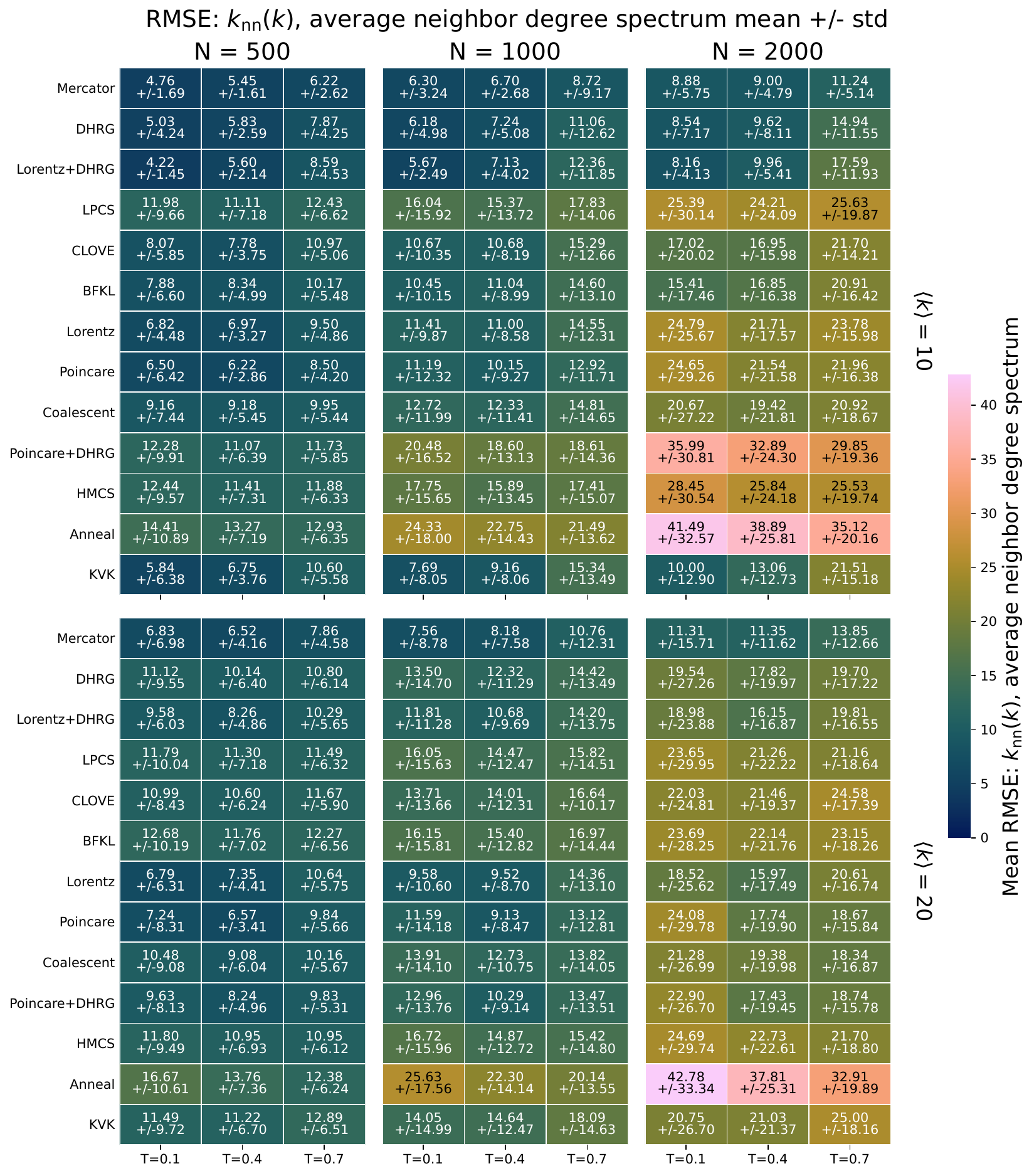} 
    \caption{Root mean square error (RMSE) between the average neighbor degree spectrum of the original graph and of the generated synthetic networks at different model temperatures. Lower values indicate closer agreement.}
    \label{fig:tr_knn_appendix}
\end{figure}

\newpage
\begin{figure}[h]
    \centering
    \includegraphics[width=0.9\linewidth]{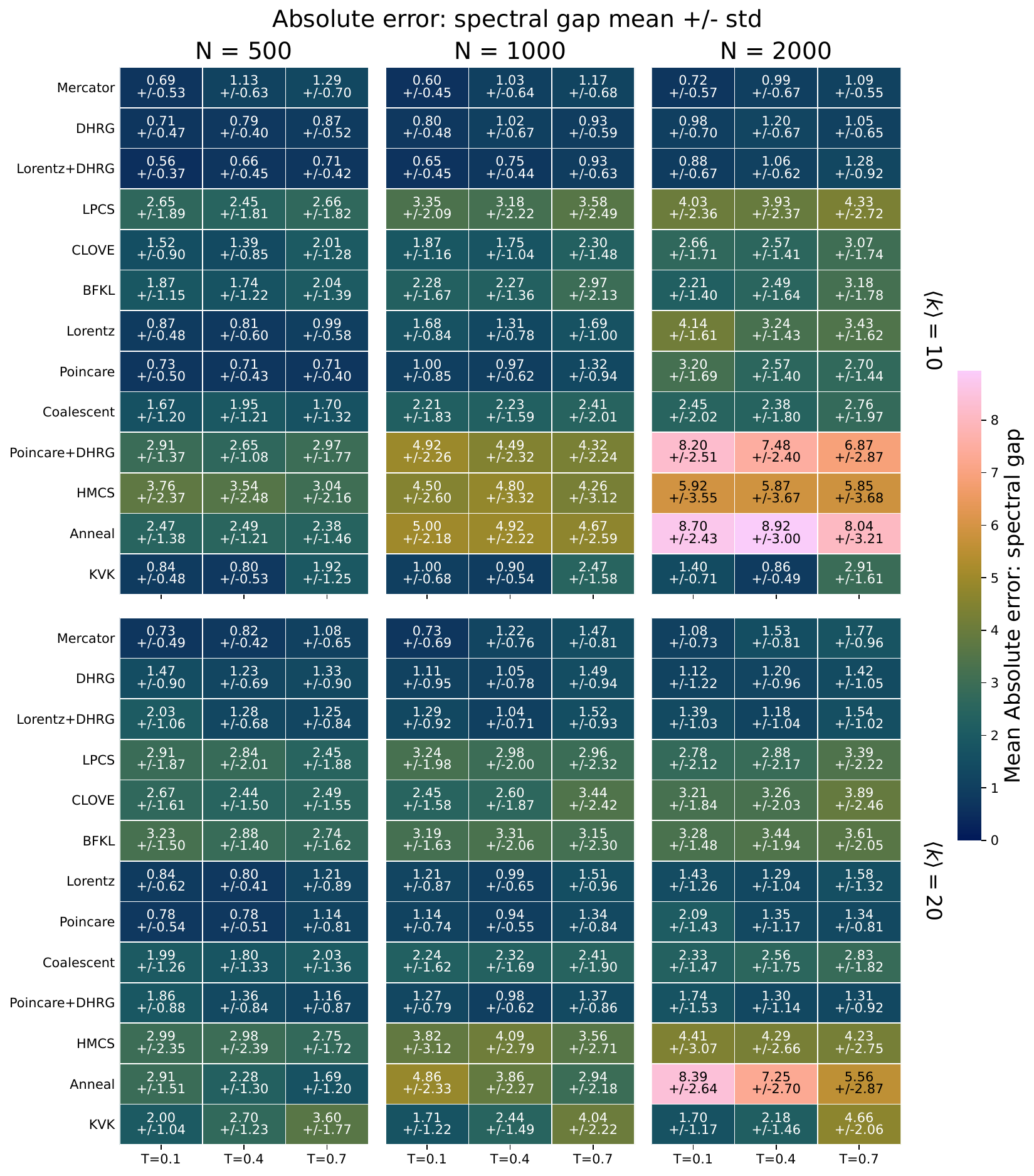} 
    \caption{Absolute error of the spectral gap of the original graph and of the generated synthetic networks at different model temperatures. Lower values indicate closer agreement.}
    \label{fig:tr_spectral_appendix}
\end{figure}

\newpage
\begin{figure}[h]
    \centering
    \includegraphics[width=0.9\linewidth]{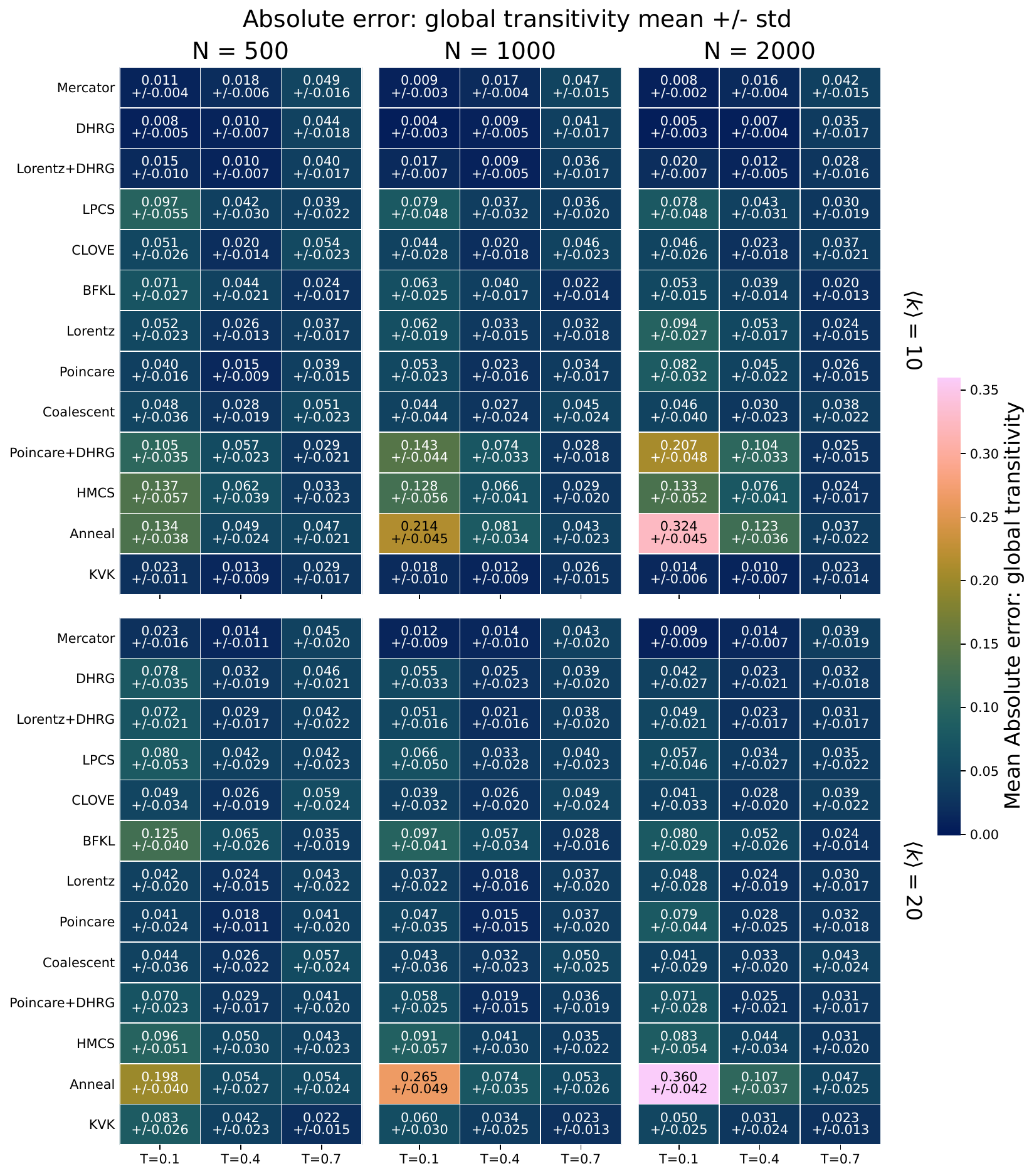}
    \caption{Absolute error of global transitivity of the original graph and those of the generated synthetic networks at different model temperatures. Lower values indicate closer agreement.}
    \label{fig:tr_transitivity_appendix}
\end{figure}

\newpage
\begin{figure}[h]
    \centering
    \includegraphics[width=0.9\linewidth]{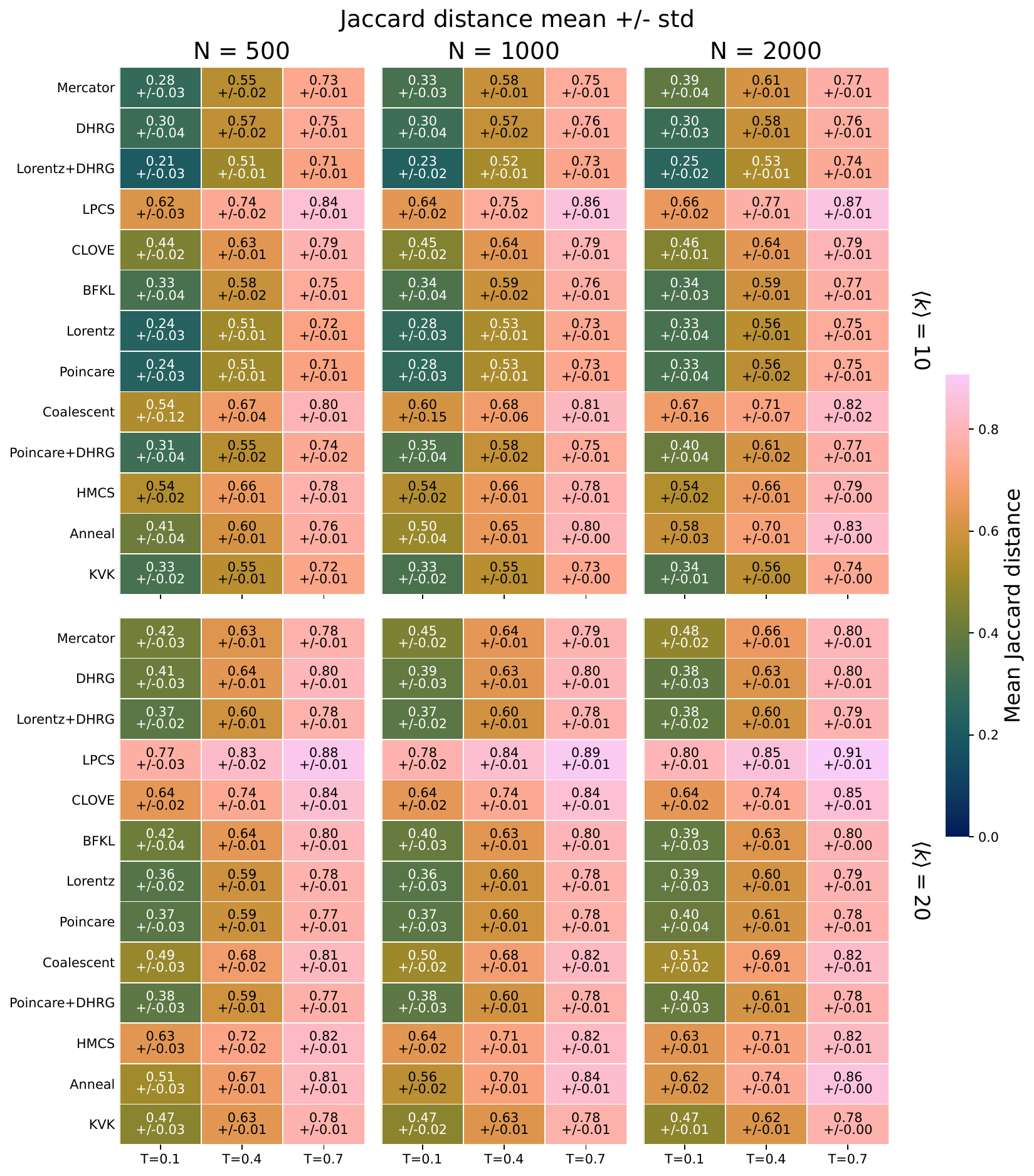}
    \caption{Jaccard distance (1 - Jaccard Similarity) between the original graph and those of the generated synthetic networks at different model temperatures. Lower values indicate closer agreement.}
    \label{fig:tr_jaccard_appendix}
\end{figure}

\newpage
\section{Additional results for the real networks}\label{apx:real_networks}

\begin{table*}[h!]
\centering
\scalebox{0.95}{
\begin{tabular}{lllcccc} \toprule
Name            & Type      & Description               &$|V|$& $|E|$ & $T$ & Source \\  \midrule
facebook        & social    & social circles        &4039 &88234 & 0.49 & \cite{snapnets} \\
openflights     & transport & flights between airports   &3397 &38460 & 0.58 & \cite{garcia2019mercator,balogh2025clove}\\
grqc            & citation  & general relativity articles    & 4158&13422 & 0.37 & \cite{snapnets} \\
yeast           & biology   & yeast metabolism      &1458 &1948  & 0.84 & \cite{jeong2001lethality} \\
diseasome       & biology   & disease relationships &516  &1188  & 0.16 & \cite{goh2007human} \\
cora & citation & machine learning papers &2708 &5429 &0.68   & \cite{sen2008collective} \\
polblogs & web & hyperlinks & 1490 & 19090 & 0.87 & \cite{adamic2005political} \\				    
CElegans        & cell      & nervous system        & 279 & 2287  & 0.73 & \cite{varshney2011structural,garcia2019mercator} \\
Drosophila1     & cell      & optic medulla         & 350 & 2887  & 0.96 & \cite{shinomiya2022neuronal} \\ 
Drosophila2     & cell      & optic medulla         &1770 & 8904  & 0.94 &  \cite{shinomiya2022neuronal} \\
ZebraFinch2     & cell      & basal ganglia (Area X)& 610 & 15342 & 0.99 & \cite{svara2022automated} \\ 
\bottomrule
\end{tabular}
}
\caption{Summary of network properties of real networks. $|V|$ -- number of nodes, $|E|$ -- number of edges, $T$ -- inferred temperature by Mercator.}
\label{tab:real_networks}
\end{table*}            

\begin{figure}[h]
    \centering
    \includegraphics[width=0.75\linewidth]{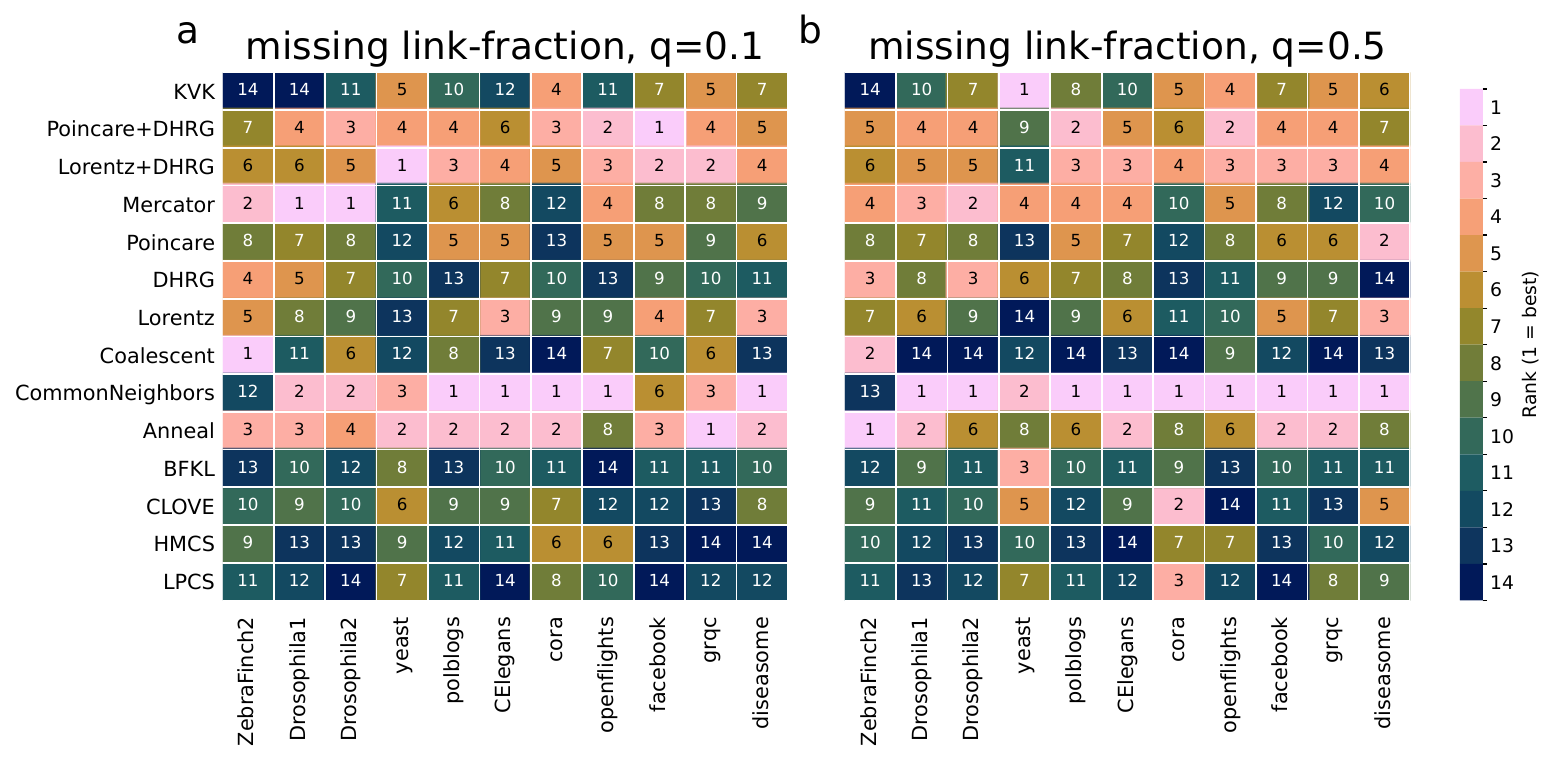}
    \caption{Ranking of the methods based on the median predictive power for each real network for missing link-fractions (\textbf{a}) $q=0.1$ and \textbf{(b)} $q=0.5$. Lower values are better ranks.}
    \label{fig:lp_real_predictive_power}
\end{figure}

\begin{figure}[h]
    \centering
    \includegraphics[width=0.75\linewidth]{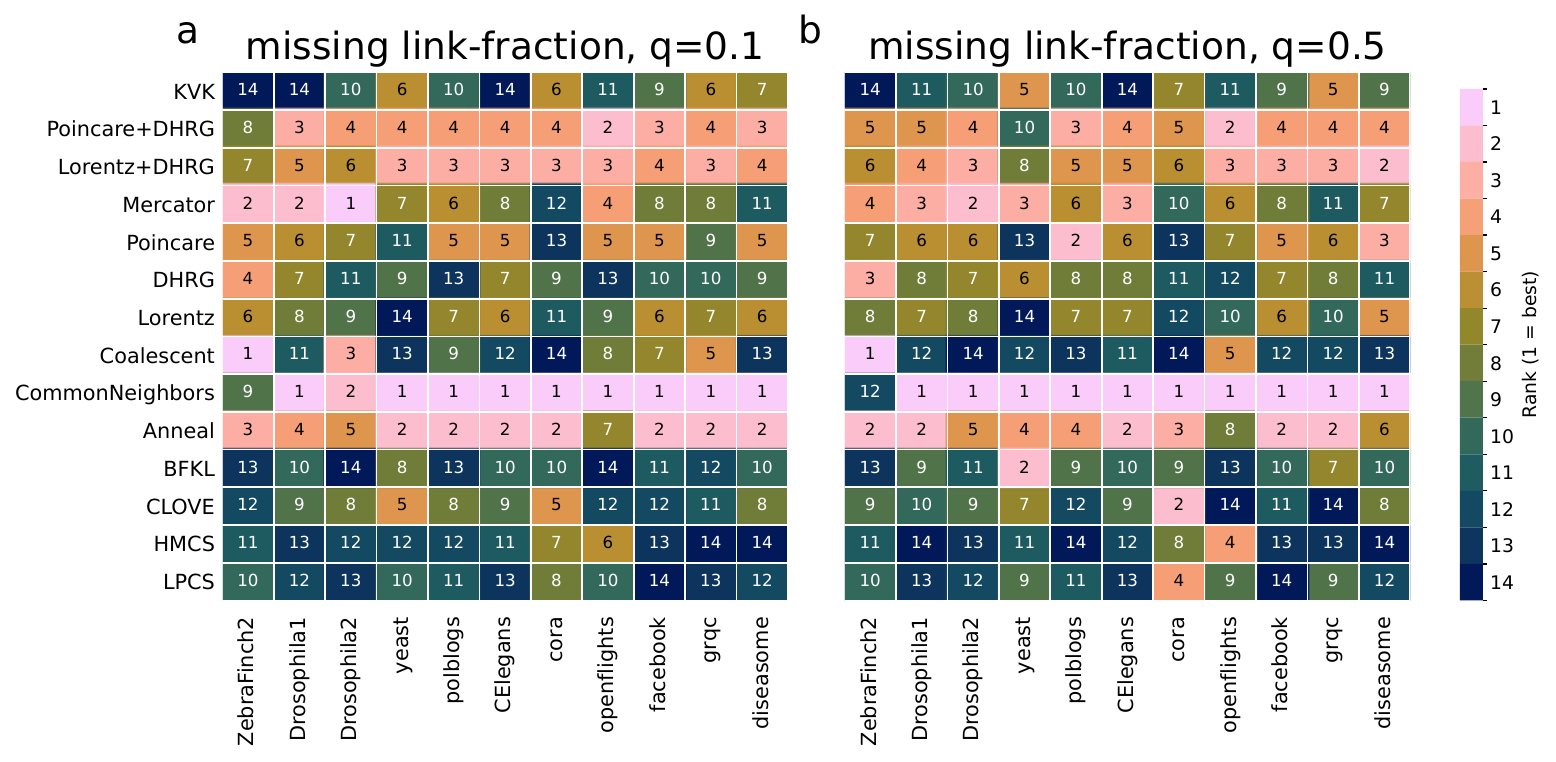}
    \caption{Ranking of the methods based on the median local precision for each real network for missing link-fractions (\textbf{a}) $q=0.1$ and \textbf{(b)} $q=0.5$. Lower values are better ranks.}
    \label{fig:lp_real_local_precision}
\end{figure}

\newpage

\begin{figure}[h]
    \centering
    \includegraphics[width=0.9\linewidth]{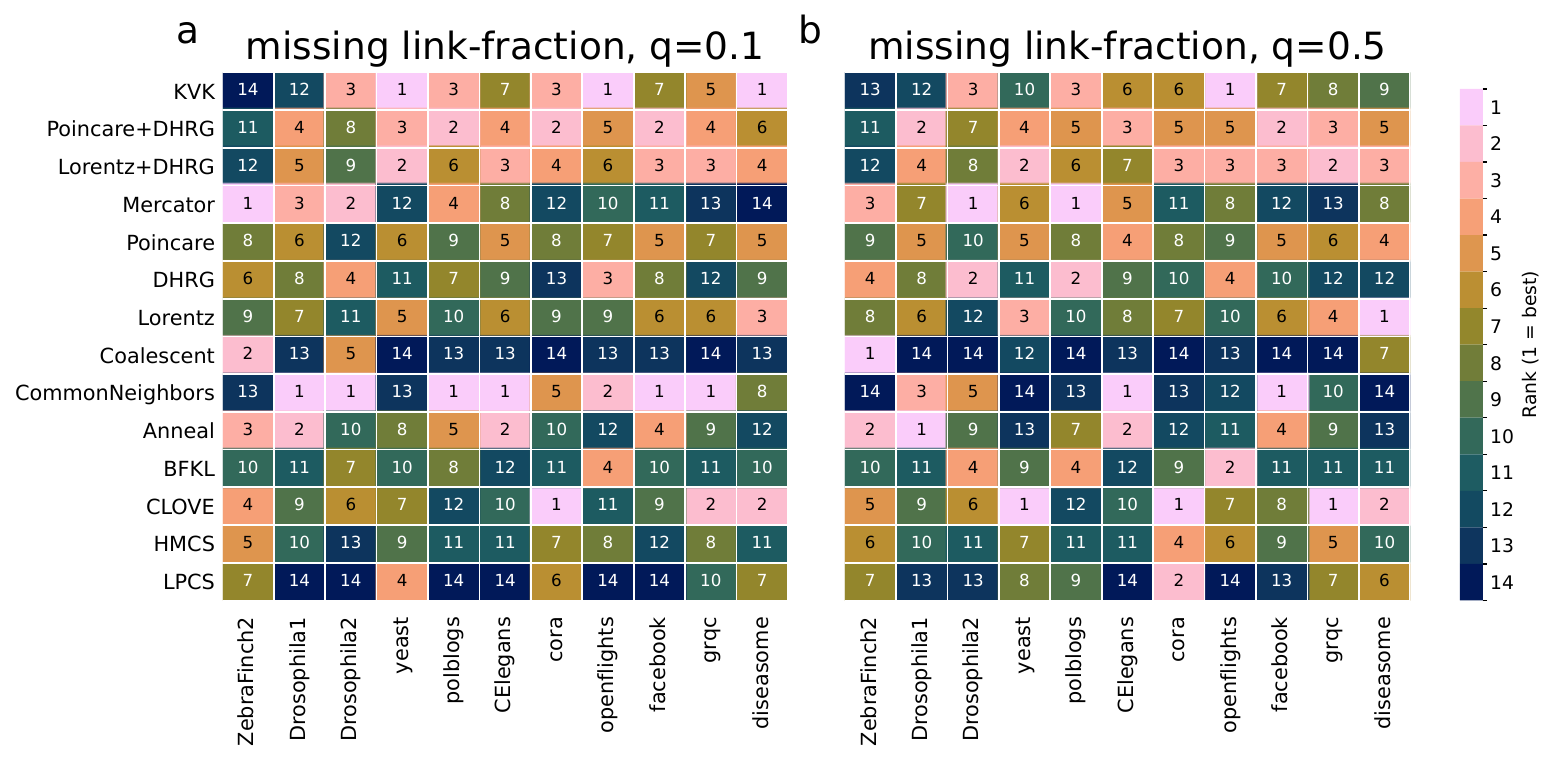}
    \caption{Ranking of the methods based on the median vertex-centric metric for each real network for missing link-fractions (\textbf{a}) $q=0.1$ and \textbf{(b)} $q=0.5$. Lower values are better ranks.}
    \label{fig:lp_real_vertex}
\end{figure}

\begin{figure}[h]
    \centering
    \includegraphics[width=\linewidth]{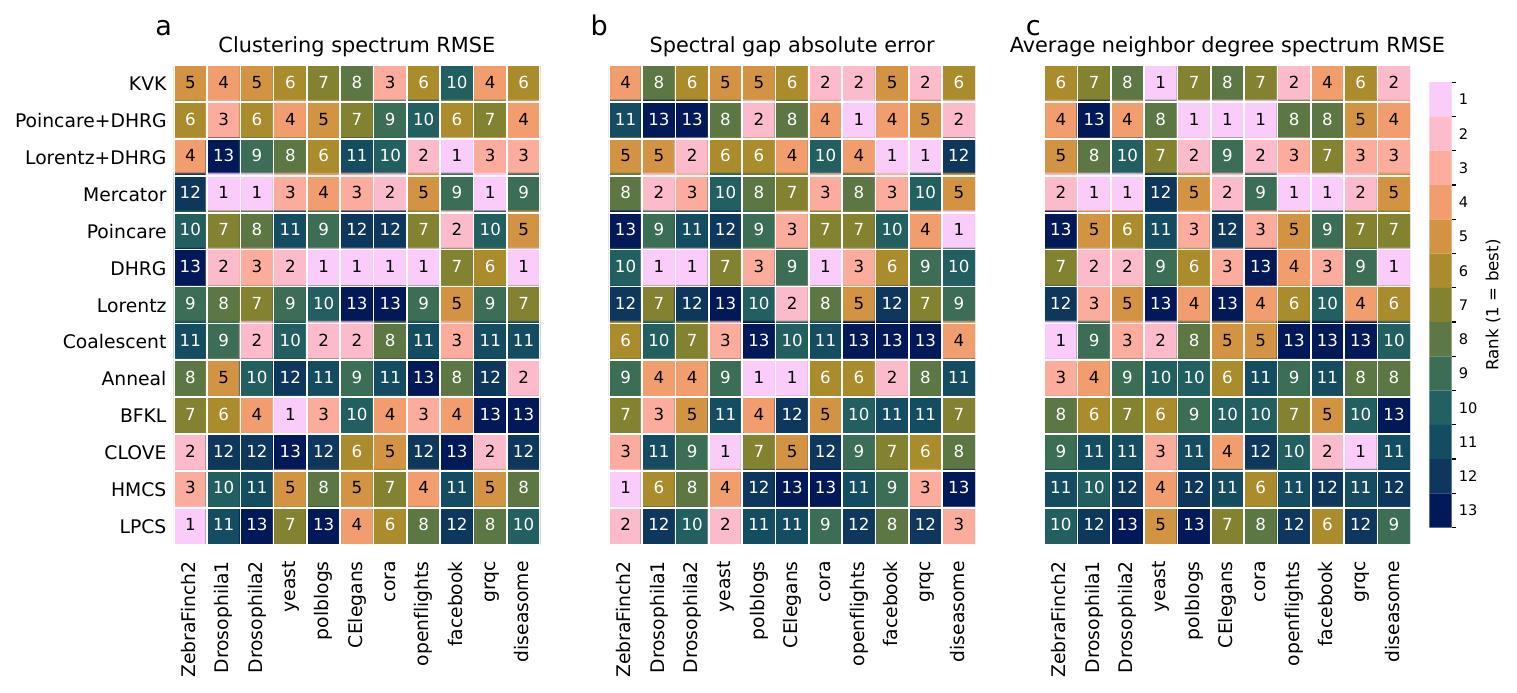}
    \caption{\textbf{Topology reconstruction in real networks.}  Rankings of the methods based on the (\textbf{a}) RMSE of clustering spectrum, \textbf{(b)} absolute error in spectral gap, and \textbf{(c)} RMSE in average neighbour degree spectrum. For each dataset, rank 1 indicates the lowest error.}
    \label{fig:tr_real_apx}
\end{figure}

\newpage

\begin{figure}[h]
    \centering
    \includegraphics[width=\linewidth]{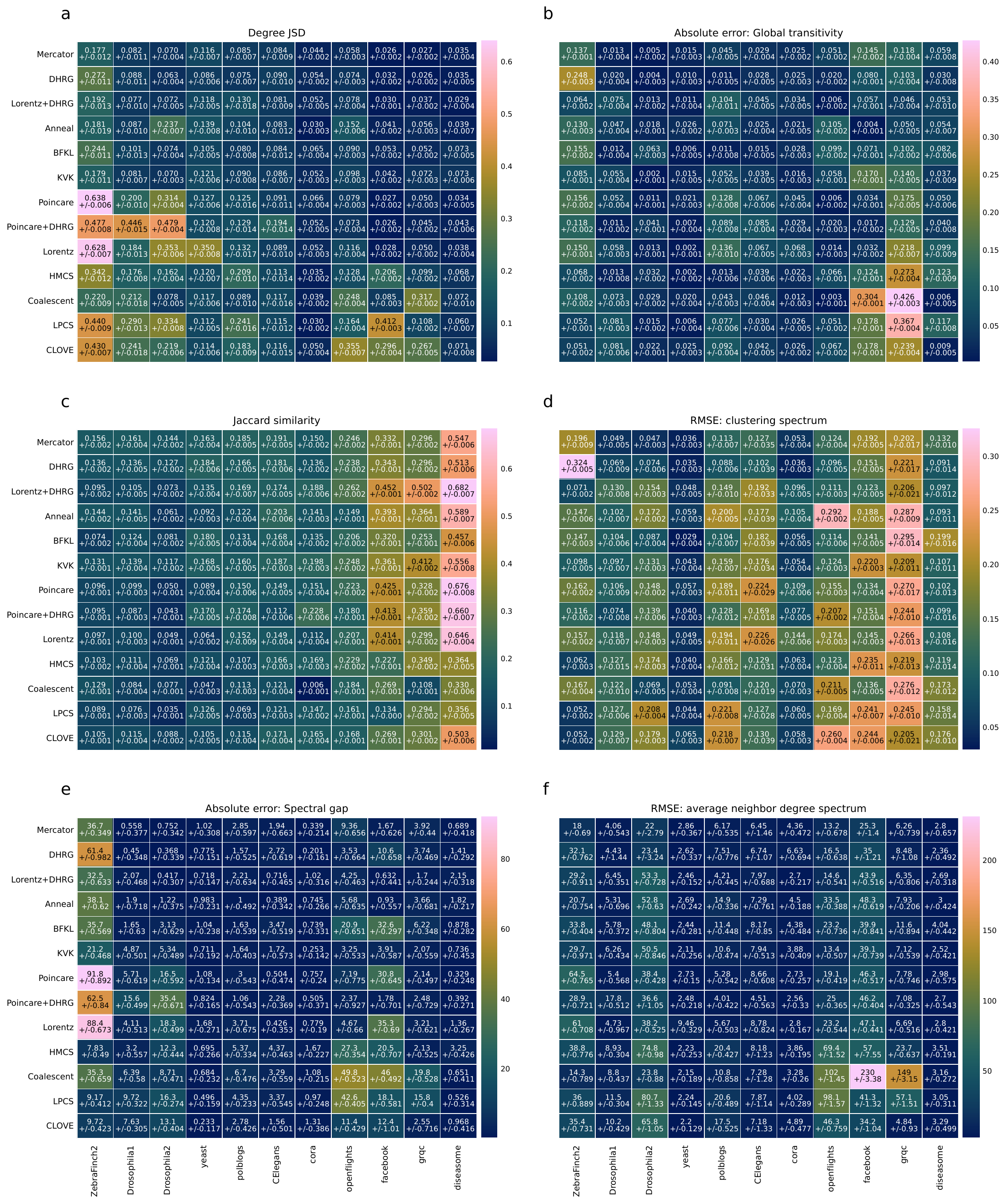}
    \caption{Topology reconstruction scores on real networks. Each heatmap compares reconstruction methods across networks, with methods on the rows and networks on the columns. Cell color encodes the mean score, and annotations report mean with standard deviation across 20 reconstruction samples. Panels show: (a) degree-distribution JSD, (b) global transitivity absolute error, (c) Jaccard similarity, (d) clustering spectrum RMSE, (e) spectral gap absolute error, and (f) average neighbor degree spectrum RMSE. Lower values are better for all error/distance measures, while higher values are better for Jaccard similarity.}
    \label{fig:tr_real_raw_values}
\end{figure}

\newpage

\begin{figure}[h]
    \centering
    \includegraphics[width=\linewidth]{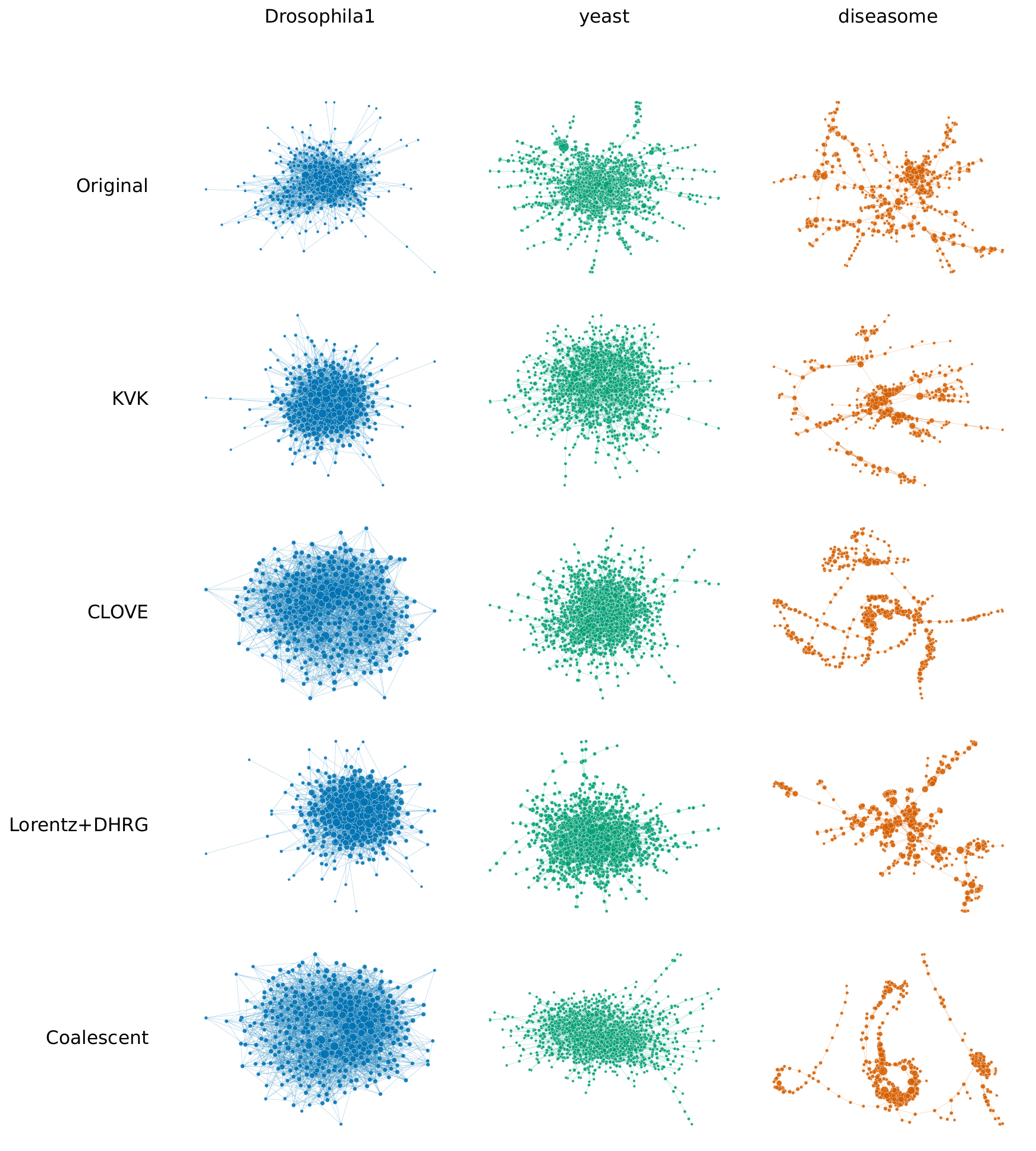}
    \caption{Visualizations of original and generated networks. Rows show methods, columns show real networks. Node size is proportional to degree.}
    \label{fig:real_networks_viz}
\end{figure}

\end{document}